\documentclass[journal]{IEEEtran}
\usepackage{color}
\usepackage{times}
\usepackage{epsfig}
\usepackage{graphicx}
\usepackage{amsmath}
\usepackage{amssymb}
\usepackage[font=small,skip=4pt]{caption}
\usepackage[font=small]{subfig}
\usepackage{tabularx}
\usepackage{multirow}
\usepackage{soul}

\usepackage[skip=5pt]{caption}
\usepackage{tabularx}
\usepackage{multirow}
\usepackage[numbers,sort&compress]{natbib}
\newcolumntype{Y}{>{\centering\arraybackslash}X}

\usepackage{color,soul}

\soulregister\cite7
\soulregister\ref7
\soulregister\pageref7

\begin{document}
	
\title{ Feature-Aligned Video Raindrop Removal with Temporal Constraints }

\author{Wending Yan$^*$,
	Lu Xu$^*$,
    Wenhan Yang, ~\IEEEmembership{Member,~IEEE}
	and Robby T. Tan,~\IEEEmembership{Member,~IEEE}
\thanks{$^*$Both authors contributed equally to this work.}
\thanks{Wending Yan and Robby T. Tan are with Yale-NUS College and Department of Electrical and Computer Engineering, National University of Singapore, Singapore (e-mail: wd.yan@nus.edu.sg, robby.tan@nus.edu.sg)}
\thanks{Lu Xu is with College of Information Science and Engineering, Northeastern University, Shenyang, China (e-mail: ian\_an\_xl@163.com)}
\thanks{Wenhan Yang is with School of Electrical and Electronic Engineering, Nanyang Technological University, Singapore (e-mail: wenhan.yang@ntu.edu.sg)}
}

\markboth{}%
{Yan \MakeLowercase{\textit{et al.}}: Feature-Aligned Video Raindrop Removal with Temporal Information}

\maketitle

\begin{abstract}
Existing adherent raindrop removal methods focus on the detection of the raindrop locations, and then use inpainting techniques or generative networks to recover the background behind raindrops. 
Yet, as adherent raindrops are diverse in sizes and appearances, the detection is challenging for both single image and video. 
Moreover, unlike rain streaks, adherent raindrops tend to cover the same area in several frames.
Addressing these problems, our method employs a two-stage video-based raindrop removal method. The first stage is the single image module, which generates initial clean results. 
The second stage is the multiple frame module, which further refines the initial results using temporal constraints, namely, by utilizing multiple input frames in our process and applying temporal consistency between adjacent output frames.
Our single image module employs a raindrop removal network to generate initial raindrop removal results, and create a mask representing the differences between the input and initial output.
Once the masks and initial results for consecutive frames are obtained, our multiple-frame module aligns the frames in both the image and feature levels and then obtains the clean background.
Our method initially employs optical flow to align the frames, and then utilizes deformable convolution layers further to achieve feature-level frame alignment.
To remove small raindrops and recover correct backgrounds, a target frame is predicted from adjacent frames. 
A series of unsupervised losses are proposed so that our second stage, which is the video raindrop removal module, can self-learn from video data without ground truths.
Experimental results on real videos demonstrate the state-of-art performance of our method both quantitatively and qualitatively.
\end{abstract}

\begin{IEEEkeywords}
Low-level vision, image enhancement and restoration, bad weather, adherent raindrop removal.
\end{IEEEkeywords}

\IEEEpeerreviewmaketitle

\section{Introduction}
Many computer vision applications assume clean videos as input. 
In outdoor scenes, however, the quality of videos can be impaired significantly by raindrops adhered to a camera lens or windscreen, since they can partially occlude the background scenes.  
While the presence of adherent raindrops can be reduced by covering the cameras or using wipers, this solution cannot be applied for every situation. In a vehicle system, there are cases that the cameras need to be installed outside the vehicles. Covering the cameras on any sides will reduce the field of view. Moreover, due to the motion of the vehicle and in combination with wind, the water drops can inevitably reach into the camera’s lens. Installing wipers is not a universal solution either, since in many cases installing a wiper on the camera lens is not possible.

Several methods have been proposed to remove raindrops. Many of these methods focus on a single image~\cite{0Attentive,2019Deep,2014Restoring,Guo2020JointRA,Jia2020ResidualAM,Luo2021WeaklySL,Hao2019LearningFS,Liang2020UsingRR}, and a few of them focus on video~\cite{porav2019can,2009Video,TANAKA2006Removal,2005Rainy,2010Realistic,2016Adherent,Alletto2019AdherentRR}.
Qian et al.~\cite{0Attentive} employs a recurrent network trained to locate raindrop regions and generates raindrop masks as spatial attention. Once the mask is obtained, it uses a generative network to inpaint the raindrop occluded regions.
While this method and other methods using single images show good results, they suffer from incorrect removal or incorrect inpainting in general, particularly for large raindrops with complex scenes behind. 
This is because, in single images, the information of the occluded background is simply non-existence in the input image.

\begin{figure}[t!]
	\begin{center}
		\vspace{-0.3cm}
		\subfloat[Input Image]{\includegraphics[width=0.235\textwidth]{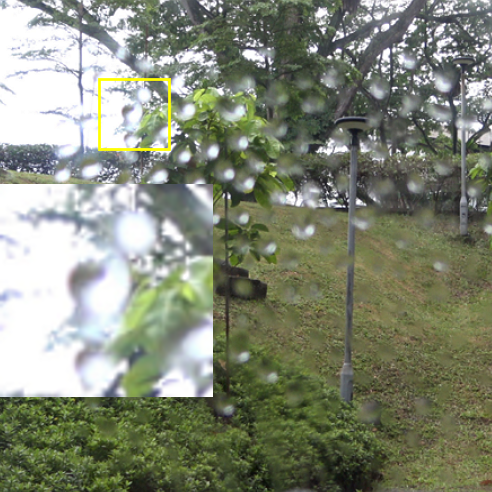}}\hfill
		\subfloat[Qian et al.~\cite{0Attentive}]{\includegraphics[width=0.235\textwidth]{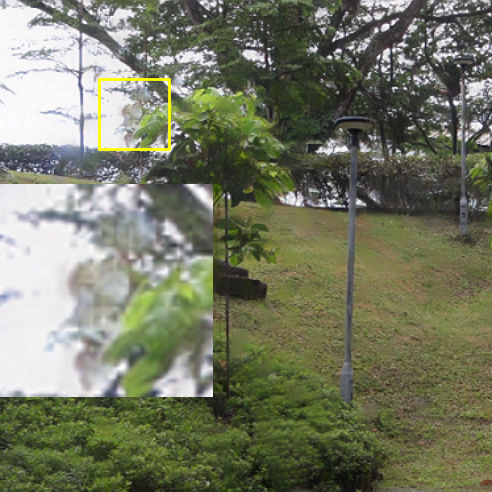}}\hfill
		\vspace{-0.2cm}
		\subfloat[Quan et al.~\cite{2019Deep}]{\includegraphics[width=0.235\textwidth]{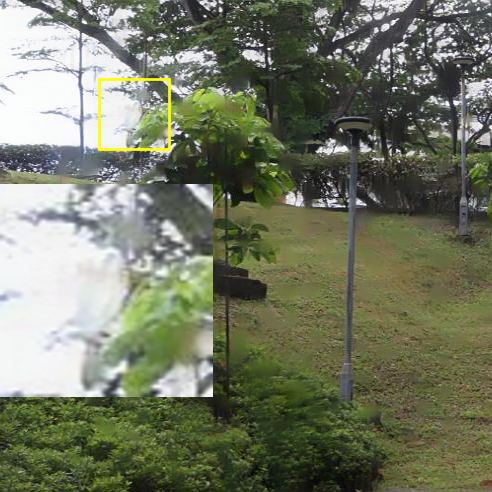}}\hfill
		\subfloat[\textbf{Our Result}]{\includegraphics[width=0.235\textwidth]{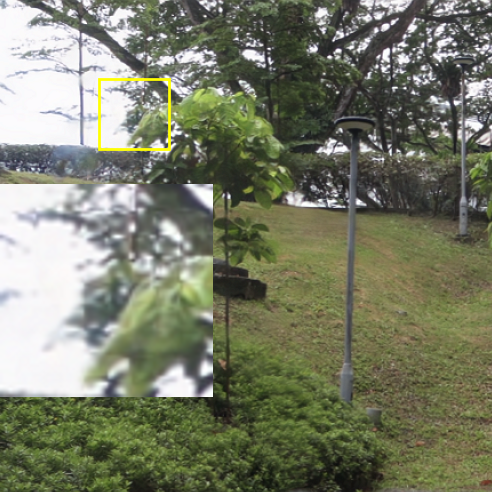}}\hfill
		\vspace{-0.2cm}
	\end{center}
	\caption{
		(a) Input image. (b) Result from Qian et al.~\cite{0Attentive}. (c) Result from Quan et al.~\cite{2019Deep}. (d) Our result. Compared to baselines, our method can recover background information behind raindrops.
	}
	\label{fig:trailer}
	\vspace{-0.1in}
\end{figure}

Unlike single-image methods, video-based methods detect/remove adherent raindrops and at the same time  maintaining	 the temporal consistency. Adjacent frames can provide information about the background covered by raindrops. Yet, in contrast to rain streaks, which distribute randomly, raindrops tend to cover the same area in several adjacent frames.  Existing single image raindrop removal methods and rain streak removal methods cannot solve this problem.

One of the recent video-based removal methods~\cite{Alletto2019AdherentRR} utilizes self-supervised attention maps to detect the raindrops.
A spatial-temporal adversarial network is then used to inpaint the occluded background, and optical flow is employed in the training stage to maintain temporal consistency. While the method works for some videos,  it is known that estimating optical flow directly from videos degraded by adherent raindrops is likely inaccurate.
Besides, the results rely on the training data, which might have some domain gaps to the testing data, causing the raindrop removal performance to drop.

In this paper,  our goal is to remove adherent raindrops from the video. 
We propose a method that employs a two-stage structure. 
The first stage is the single image module, which generates initial clean results, and masks representing the differences between the input frames and the initial clean results.
The second stage is the multiple frame module, which further refines the result by imposing our temporal constraints.

Our single image module is trained in a fully supervised manner using the pairs of raindrop images and the corresponding clean images. An attention-guided adversarial network is built to predict the results. A raindrop mask is defined as the absolute distance between a raindrop image and a predicted background image.
In the second stage, our multi-frame module aligns the initial results in the image level using optical flow, and in the feature level using deformable convolution layers.  The aligned adjacent frames are then used to predict the clean current frame.
In this work, we have designed a group of unsupervised losses, which make our video raindrop removal module be self-supervised.
These unsupervised losses are important in our training stage to reduce the domain gaps between the synthetic data and the real-world data.

Fig.\ref{fig:trailer} shows our result. In the input frame, the leaves in the yellow box are covered by a few raindrops. The baseline method~\cite{0Attentive} in (b) simply replaces raindrops with light green patches, whose green color is borrowed from the wrong leaves in previous frames.  The other baseline~\cite{2019Deep} in (c) mistakenly fills the central raindrop with the sky color. None of the baseline methods can correctly recover the correct color and texture details of the leaves. 

In summary, our contribution are as follow:
\begin{itemize}	
	\item
	We introduce a video-based raindrop removal method that can recover the background details robustly from adjacent frames. To our knowledge, it is the first attempt to include image-level and feature-level alignment and to make use of the adjacent information to recover the details without using an inpainting technique or a generative network to recover the background behind raindrops.
	
	\item 
	We propose a two-stage method, where the first stage provides raindrop masks as well as initial results, and the second stage refines the results.
	In the second stage, our frame alignment is jointly optimized with the raindrop removal. Moreover, unlike existing methods, instead of directly estimating optical flow from the input frames, our two-stage strategy can estimate the optical flow and remove raindrops more accurately. 
	
	\item 
	We introduce a few unsupervised losses, such that our method can learn from real-world data. Our unsupervised losses integrate with both temporal correlation and consistency, such that our method can be self-supervised to minimize the domain gap in the video raindrop removal module. To the best of our knowledge, this is the first attempt at video raindrop removal in a self-supervised manner.
\end{itemize}
In this paper, we assume that the background is moving faster than raindrops, which is a reasonable assumption since, in many real-world applications, the camera is moving.

\section{Related Work}
Rain may cause different kinds of degradation in images and videos such as rain streaks, raindrops, poor visibility, and background blur. Recently, many works have been done to restore a clean image from rainy weather. 
Most of these works focus on rain streak removal~\cite {2012Automatic,2015Removing,2015Exploiting,2008Using,2006Rain,2020Self}.
Different from rain streaks, raindrops usually cover a much bigger area and have more irregular geometric structures~\cite{2016Adherent}. Thus, raindrop removal can be very challenging. 
The existing raindrop removal methods can be roughly categorized as single-image methods~\cite{2014Restoring,0Attentive,2019Deep} and multi-images methods~\cite{2009Video,TANAKA2006Removal,2005Rainy,2010Realistic,2016Adherent}.

Single image raindrop removal methods only take one degraded image as input and restore the background. With the development of the learning-based method, single image raindrop removal has grown rapidly. 
Pix2Pix~\cite{Isola2017ImagetoImageTW} and CycleGAN~\cite{Zhu2017UnpairedIT} are commonly used to deal with such kinds of image or video enhancement tasks. However, to remove raindrops while maintaining the background, a specially designed strategy is crucial.
Eigen et al.~\cite{2014Restoring} was the very first one to use deep learning for raindrop removal, and with a shallow convolutional network, the clean background can be restored.
Porav et al.~\cite{porav2019can} introduce a novel stereo system to generate real-world raindrop data with corresponding clean images.A deraining network architecture is proposed based on Pix2PixHD~\cite{Wang2018HighResolutionIS}. Motivated by the fact that much of the structure of the input image should be kept, the addition of additive skip connections is used. These methods follow the idea to first detect the raindrop and then restore the back scene. However, single image raindrop removal methods rely on generating networks for inpainting. The recovered result may suffer artifacts and fakeness.
Qian et al.~\cite{0Attentive} proposed an attentive GAN to detect the location of raindrops and recover the background. A recurrent network is used to detect the raindrops and generate corresponding attention maps. Then the generated map is used with a CNN and a discriminator to restore the latent image under the mask. Paired single with raindrop image and clean image is used for training. For training data, the ground-truth location mask is obtained by submitting the raindrop images and the clean images with some morphological operations for post-processing.

Based on Qian et al.~\cite{0Attentive}’s data, Quan et al.~\cite{2019Deep} proposed a shape-driven attention mechanism and channel re-calibration to guide the CNN. The proposed shape attention exploits the physical prior of raindrop, convexness, and contour closedness, to locate the raindrops. Channel re-calibration strategy is used to boost the robustness of various raindrop types. 

Some work has been done in video raindrop removal. Yamashit et al.~\cite{2005Rainy} uses multiple images to jointly eliminate raindrops. 
Methods such as PCA and bezier curve fitting are used to recognize raindrops~\cite{2005Rainy,2010Realistic}.
Roser et al.~\cite{2009Video} detect raindrops using a photometric model. 
To restore the clean background, information from adjacent frames is fused. You et al.~\cite{2016Adherent} use the temporal change intensity of raindrop pixels and non-raindrop pixels to detect the location of raindrops. 
Video completion technique and temporal intensity derivative are used to handle completely occluded region and partially occluded region. 
Alletto et al.~\cite{Alletto2019AdherentRR} is the first one to use deep-learning to deal with the video raindrop removal task. A self-supervised attention map and a spatio-temporal generative adversarial network is proposed to solve the problem. A single image raindrop removal model is first trained. Then a spatio-temporal generator is proposed and jointly trained with optical flow.

Recently, learning-based methods achieved great results on video inpainting. The main idea of these methods is to use information from adjacent frames to help with the inpainting in the current frame. 
Methods like ~\cite{oh2019onion,li2020short} use attention mechanisms to retrieve information from adjacent frames. 
Benefit from the development of optical flow estimation~\cite{sun2018pwc,Ilg2017FlowNet2E,2016FlowNet}, method like~\cite{Xu2019DeepFV,Gao2020FlowedgeGV} proposed flow guided video inpainting strategy. 
These methods rely on accurate optical flow. Another commonly used way is 3D convolution~\cite {Chang2019FreeFormVI,Hu2020ProposalBasedVC}. 
In this paper, we take advantage of both flow-guided methods and 3D convolution and propose a self-learned model to refine the results.

\begin{figure*}[t!]
	\begin{center}
		{\includegraphics[width=\textwidth]{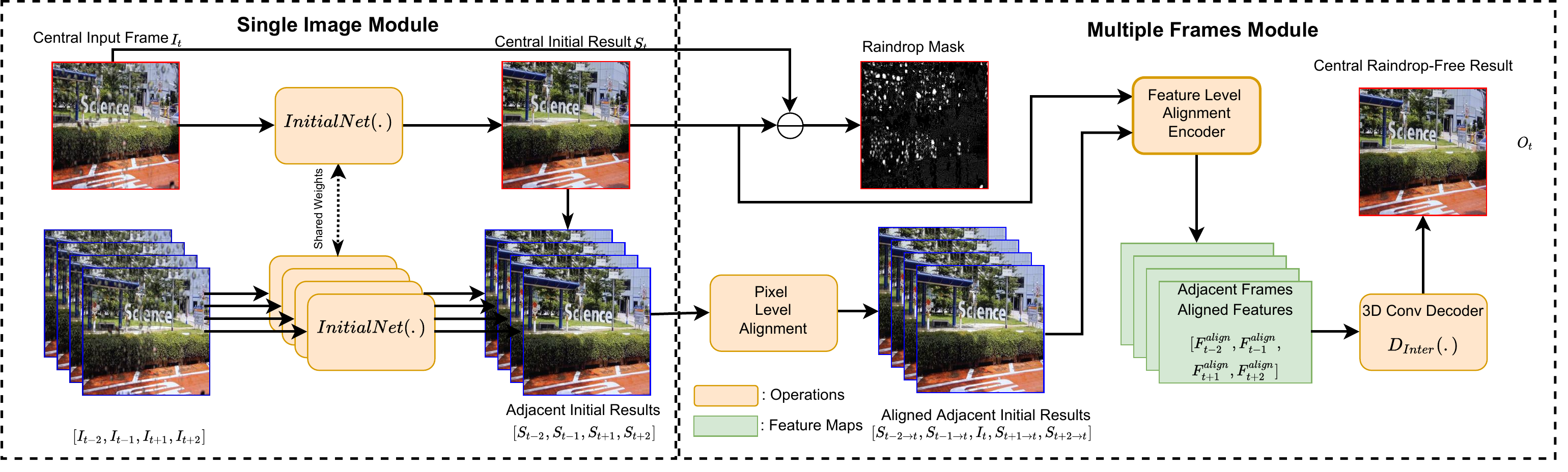}}\hfill
	\end{center}
	\caption{
		The pipeline of our framework, which consists of two components: Single Image Module (InitialNet) and Multiple Frames Module (VideoNet).
	}
	\label{fig:pipeline}
\end{figure*}

\section{Proposed Method}

Taking $N$ raindrop frames as input, e.g.,  $\left ( I_{t-2},I_{t-1},I_{t},I_{t+1},I_{t+2} \right )$ when we set $N=5$, our goal is to restore  $J_{t}$, which is the clean background of $I_{t}$, the current frame. 
In the first stage, an initial frame and corresponding mask are obtained by a single image raindrop removal. 
In the second stage, we freeze the first stage and take advantage of the temporal constraints to refine the raindrop removal results. 

\subsection{Initial Stage}
Fully supervised methods like Qian et al.~\cite{0Attentive} or Quan et al.~\cite{2019Deep} use single images for raindrop detection and removal.
While the results are not completely accurate, particularly for small raindrops due to the inaccurate detection and relatively large raindrops due to the inpainting process, several raindrops are removed and recovered, at least better than the original degraded input. For this reason,  our second stage, which is the multi-frame stage, will use the output of this initial stage, instead of the original input frames.

Fig.\ref{fig:pipeline} shows our pipeline, where in the initial stage, we will have the initial results of some adjacent input frames.
Taking one frame with raindrop $I_{t}$ as input, an initial result $S_{t}$ can be restored:
$S_{t} = InitialNet(I_{t})$,
where $InitialNet(\cdot)$ is a fully supervised single image raindrop removal method. 
With the initial results, the corresponding masks that each indicate the location of the raindrops, can be obtained by subtracting the raindrop frame  $I_{t}$ and  $S_{t}$:
\begin{equation}
	M_{t}= \left | I_{t}-S_{t} \right |,
\end{equation}
Besides $M_{t}$, the initial result $S_{t}$ provides better quality of the background (than the original input), and thus benefits our frame alignment process in our next stage. Inspired by Qian et al~\cite{0Attentive}, we employ an auto-encoder with an attention mechanism as our model. The model is trained on real raindrop data~\cite{0Attentive} in a fully supervised way.

\subsection{Multi-Frames Stage}
\begin{figure}
	\begin{center}
		{\includegraphics[width=\columnwidth]{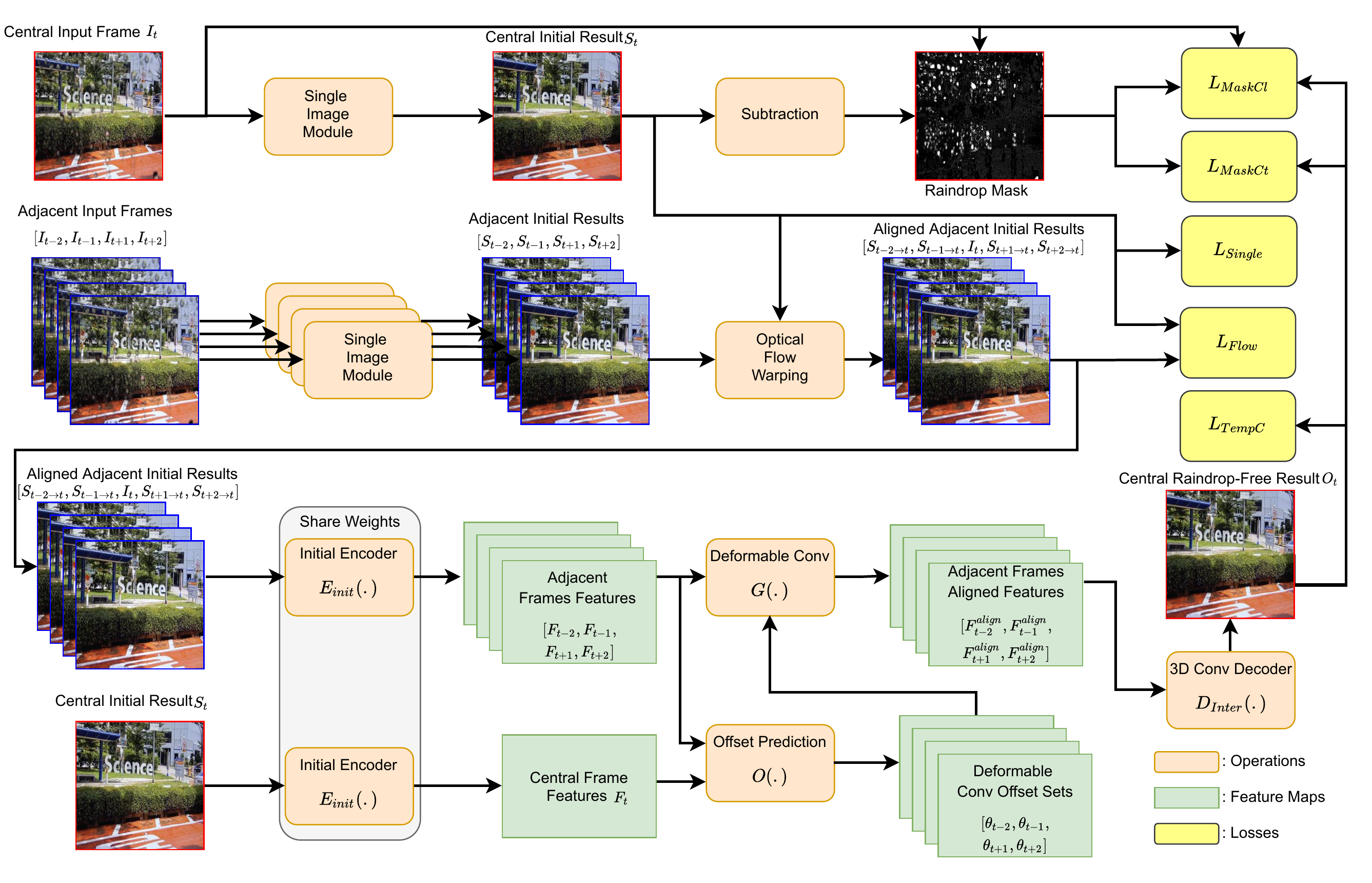}}\hfill
	\end{center}
	\caption{
		The architecture of our Multiple Frames Module (VideoNet).
	}
	\label{fig:stage2}
\end{figure}

As shown in our pipeline in Fig.\ref{fig:pipeline}, once the masks and initial results are obtained, we then refine the background recovery using the temporal consistency and correlation with adjacent frames.
The basic idea is that we align the video frames in the image level (the pixel level alignment using optical flow) and feature level (the feature level alignment using the deformable convolution encoder), and then estimate the clean background (the 3D conv decoder). A few unsupervised losses are proposed to make our network self-learned in this stage.
Fig.\ref{fig:stage2} shows the detailed processes of our method. There are 3 main processes in this multi-frame stage: optical flow, self-alignment encoder, and raindrop removal decoder.

\vspace{0.4cm}

\noindent \textbf{Optical Flow} 
Using the initial raindrop removal results, $S_t$'s, we compute the optical flow and warp to align the raindrop video frames.
A pretrained FlowNet~\cite{ilg2017flownet} is employed to estimate the optical flow from the initial results $S_{i}$ and $I_{j}$, expressed as:
\begin{equation}
	F_{i\rightarrow j}=FlowNet\left ( S_{i},I_{j} \right ),
\end{equation}
where $i\rightarrow j$ denotes the flow from the $i$-th frame to $j$-th frame. 
Then, we warp the initial results to the target time step:
\begin{equation}
	S_{i\rightarrow j}=W\left ( S_{i},F_{i\rightarrow j} \right ).
\end{equation}
To further improve the estimation accuracy, we finetune the pretrained network with the initial raindrop removal results. 
The warped frames should now be aligned to the current one which can be expressed as:
\begin{equation}
	L_{flow}=\sum_{i=t-s}^{t-1}\left \| S_{i\rightarrow t} -I_{t}\right \|{_{2}^{2}}+\sum_{i=t+1}^{t+s}\left \| S_{i\rightarrow t} -I_{t}\right \|{_{2}^{2}}.
\end{equation}
where $L_{flow}$ is expected to be considerably low if the alignments are correct.

\begin{figure*}[t!]
	\begin{center}
		\captionsetup[subfigure]{labelformat=empty}

		\subfloat[Input Image]{\includegraphics[width=0.25\textwidth]{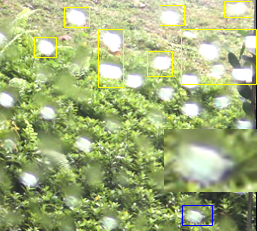}}\hfill	
		\subfloat[Initial Result]{\includegraphics[width=0.25\textwidth]{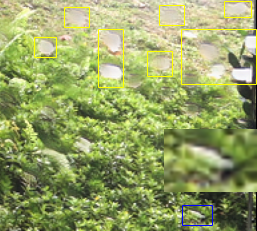}}\hfill
		\subfloat[Mask]{\includegraphics[width=0.25\textwidth]{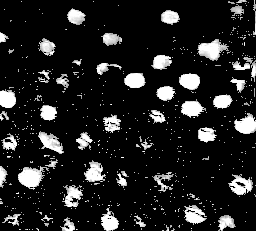}}\hfill
		\subfloat[Final Result]{\includegraphics[width=0.25\textwidth]{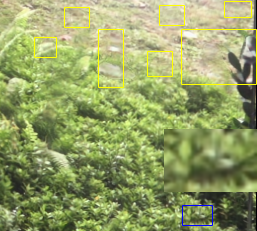}}\hfill\\
		
		\subfloat[Input Image]{\includegraphics[width=0.25\textwidth]{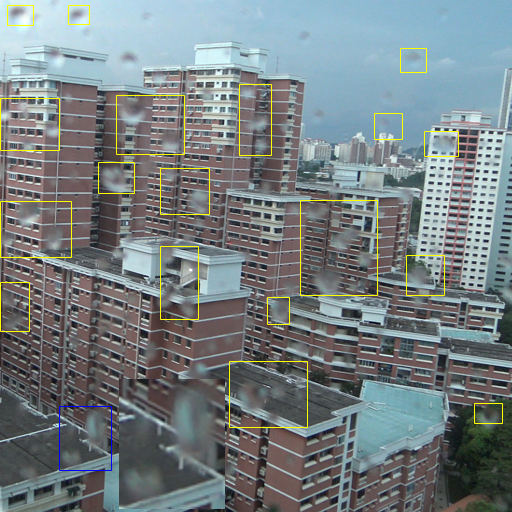}}\hfill	
		\subfloat[Initial Result]{\includegraphics[width=0.25\textwidth]{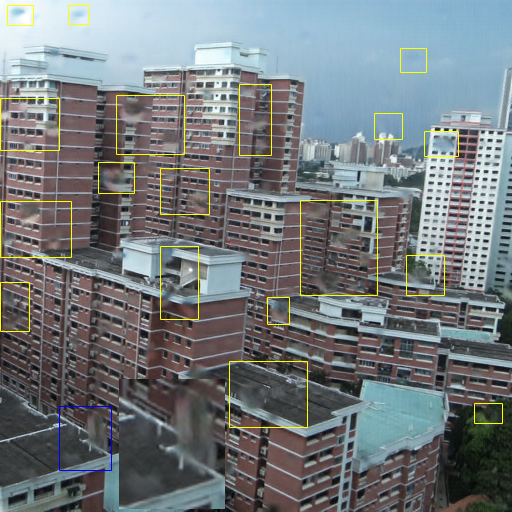}}\hfill
		\subfloat[Mask]{\includegraphics[width=0.25\textwidth]{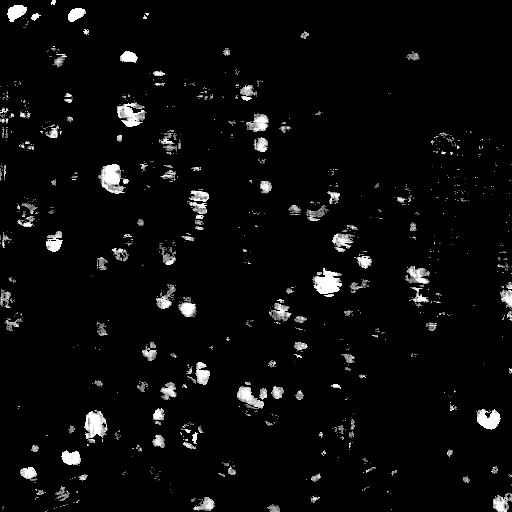}}\hfill
		\subfloat[Final Result]{\includegraphics[width=0.25\textwidth]{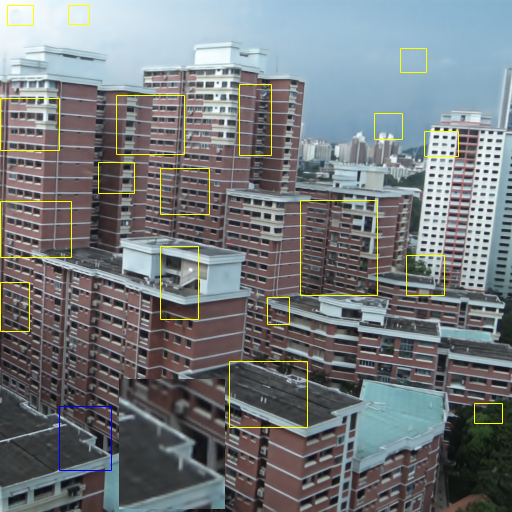}}\hfill\\	
		
	\end{center}
	\caption{
		Our $InitialNet$ provides the initial result and mask of raindrop from the single input image. These initial results and masks benefit estimating our final result in our Multiple Frames Module. We show the areas indicated by the blue boxes in larger scale. Zoom in the areas indicated by the yellow boxes to see the the effectiveness of our further refinement.
	}
	\label{mask}
\end{figure*} 

\vspace{0.4cm}

\noindent \textbf{Self-Alignment Encoder}
Having done the image level alignment, we align the input frames at the feature level. To achieve this, we employ deformable convolution layers~\cite{dai2017deformable}, 
which can extract similar features from several input objects, even if these objects are located at different spatial locations in the input frames. Hence, despite the different locations of the input features, the extracted features by the deformable convolution network are aligned.

Unlike the common convolutional layers with a fixed kernel configuration, the kernel of this convolution has grids similar to the dilated convolution layers. Such grids contain a group of spatial offsets for kernel convolution. But unlike the dilated convolution layers that have fixed offsets, the offsets in these deformable convolution layers are learnable. 
With proper training on these offsets, one can further align the frames in the feature domain.

Taking the aligned frames and the current frame, i.e., $\left ( S_{t-2\rightarrow t},S_{t-1\rightarrow t},I_{t},S_{t+1\rightarrow t},S_{t+2\rightarrow t} \right )$ as input, an initial encoder $E_{init}(\cdot)$ is used to extract feature maps:
\begin{equation}
	F_{i}=E_{init}\left ( S_{i} \right ),
\end{equation}
\begin{equation}
	F_{t}=E_{init}\left ( I_{t} \right ),
\end{equation}
where $i\in \left \{ t-2,t-1,t+1,t+2 \right \}$. 
To further align the frames at feature level, we predict the offsets $\theta {_i}$, which are  predicted with respect to the central feature map $F_{t}$:
\begin{equation}
	\theta {_i} = O\left (F_{i},F_{t}  \right ),
\end{equation}
where $O$ is the function for offsets prediction, which is designed as few convolutional layers.
Once the offsets $\left ( \theta _{t-2},\theta _{t-1},\theta _{t+1},\theta _{t+2} \right )$ are obtained, the aligned feature maps can be extracted using the deformation convolution layers:
\begin{equation}
	F_{i}^{align}=G\left ( F_{i},\theta _{i} \right ),
\end{equation}
where $i\in \left \{ t-2,t-1,t+1,t+2 \right \}$ and  $G$ is the function of the deformation convolution layers, and $i\in \left \{ t-2,t-1,t+1,t+2 \right \}$.

\vspace{0.4cm}

\noindent \textbf{Raindrop Removal Decoder}
Once the adjacent frames are aligned, their feature maps produced by the self-aligned encoder should contain the same texture information, which represents the background scenes. These feature maps are used by our decoder to generate clean background frames.

Let $D_{inter}(\cdot)$  denote the decoder.
To further make use of the temporal feature information, a 3D convolutional layer is used to predict raindrop free output $O_{t}$ from a set of aligned feature maps$\left ( F_{t-2}^{aligned},F_{t-1}^{aligned},F_{t+1}^{aligned},F_{t+2}^{aligned} \right )$. The decoder is encouraged to perform interpolation based on the adjacent frames features.
Taking the camera as a reference, adherent raindrops and the clean background appearance in the input video have different moving speeds, since the raindrops are much closer to the camera. Due to this relative motion between the raindrops and background, most of the background information of the current frame can be obtained from these features, including regions covered by raindrops.

In real-world scenes, raindrops may be small and hard to be detected and were undetected in our initial stage. These raindrops might cover only several pixels. Thus, even a tiny displacement between the raindrop and the camera will completely change their positions in the captured frames and make these small raindrops distribute randomly instead of appearing at the same location in adjacent frames.  Thus, with an interpolation strategy, our model can automatically remove these missed raindrops. 
This process can be expressed as:
\begin{equation}
	O_{t} = D_{inter} \left ( F_{t-2}^{aligned},F_{t-1}^{aligned},F_{t+1}^{aligned},F_{t+2}^{aligned} \right ).
\end{equation}
In raindrop videos, since the motions of raindrops and the background are different, the current frame is likely to have the missing background information presented in the adjacent frames. Thus, our decoder is trained based on the temporal correlation to guarantee that the needed information flows from the adjacent frames to the current frame.
Figs.~\ref{mask} shows the results from each stage of our method. Taking one raindrop frame as input, the initial result and mask are obtained using the single image module.
The generated clean background is further refined using our multiple frames module that not only removes the residual small raindrops but also boosts the removal performance.

\subsection{Loss Functions}

\noindent \textbf{Single Image Loss} 
Our single image raindrop removal is fully supervised. Inspired by existing singe image raindrop removal methods~\cite{0Attentive,2019Deep}, we apply a pixel level loss, a perceptual loss and a generative adversarial loss here, which can be expressed as:
\begin{equation}
	\mathcal{L}_{single} \! = \! \left \| \! I_{t} \! - \! S_{t} \! \right \|_{2}^{2} +\left \| \! VGG(I_{t}) \! - \! VGG\left ( S_{t} \! \right )\right \|_{2}^{2} +log(1- \! D\left ( S_{t} \right )),
\end{equation}
where $VGG(\cdot)$ is the feature maps extracted by the pretrained model and $D(\cdot)$ is the discriminator. 
During the training, we freeze the single image raindrop removal module and train our refinement network. 
Once we give the initial results and raindrop masks, our subsequent network can be trained in a self-learning way.

\vspace{0.4cm}

\noindent \textbf{Mask Consistency Loss} 
Adherent raindrops cover some parts of the scene. To make sure the network learns the information from the non-raindrop area, and to regularize the network to process different regions adaptively, a mask consistency loss is adopted. The loss enforces that the background of the input image remains consistent:
\begin{equation}
	\mathcal{L}_{mask}^{ct} = 
	\left\| Mask_t (O_{t}-I_{t} ) \right\|_{2}^{2},
\end{equation}
where $Mask_t$ is the function of the mask filtering, which ensures that only the pixels outside the raindrops are taken into consideration.

\vspace{0.4cm}

\noindent \textbf{Mask Correlation Loss}
In our method, we assume the background is moving faster than raindrops, which 
is a reasonable assumption in many real-world applications where the camera is moving. Hence, raindrops adhered to a lens have relative motion with the background taken by the camera. The difference in the motion patterns between raindrops and backgrounds makes it possible to use the information from adjacent frames to recover the background in the current frame.
Thus, we proposed a mask correlation loss that requires the estimated output frame to be consistent with all the aligned adjacent video frames on the non-raindrop area. The loss is expressed as:

\begin{equation}
	\mathcal{L}_{Mask}^{cl} = \sum_{i}^{}\frac{1}{n}\left \| Mask_{i}\left ( \tilde{O_{i}}-I_{i}\right ) \right \|_{2}^{2},
\end{equation}
\begin{equation}
	\tilde{O}_{i} = W(O_{t},F_{t\rightarrow i}),
\end{equation}

where $i\in \left \{ t-2,t-1,t+1,t+2 \right \}$. Variable $n$ is the number of the adjacent frame, which is 4 in our case.
$O_{t}$ is the current frame after raindrop removal.
$W$ is the function of the warping operation. 

\vspace{0.4cm}

\noindent \textbf{Temporal Consistency Loss}
Since our method is video-based, we further explore making use of the information of adjacent frames by designing a temporal consistency loss.
Our temporal consistency loss consists of two parts.
The first one $\mathcal{L}_{flow}$ is used to refine the estimation of optical flow. We finetune the pretrained network by forcing the warped frames to be aligned to the current one.
The second one requires the estimated output frame consistent to all the aligned adjacent video frames. This can further refine the optical flow and  provide consistent inpainting results. We defined temporal consistency loss as:
\begin{equation}
	\mathcal{L}_{flow} = \sum_{i}^{}\frac{1}{n}\left \| Mask_{i}\left ( \tilde{S_{i}}-S_{i}\right ) \right \|_{2}^{2},
\end{equation}
\begin{equation}
	\mathcal{L}_{temp} = \sum_{i}^{}\frac{1}{n}\left \| Mask_{i}\left ( \tilde{O_{i}}-O_{i}\right ) \right \|_{2}^{2},
\end{equation}
\begin{equation}
	\tilde{S}_{i} = W(S_{t},F_{t\rightarrow i}).
\end{equation}
\begin{equation}
	\tilde{O}_{i} = W(O_{t},F_{t\rightarrow i}).
\end{equation}
where $i\in \left \{ t-2,t-1,t+1,t+2 \right \}$. $O_{t}$ is the raindrop removal results of the current frame.

Overall, our loss in the second stage can be expressed as: 
\begin{equation}
	\mathcal{L}_{all}= \mathcal{L}_{flow}+\mathcal{L}_{mask}^{ct}+\mathcal{L}_{mask}^{cl}+
	\lambda_{t}\mathcal{L}_{temp}.
\end{equation}
Our mask consistency loss $\mathcal{L}_{mask}^{ct}$ and mask correlation loss $\mathcal{L}_{mask}^{cl}$ are both $L2$ losses between two images, and thus have the same range of values. Hence, we set the same weight for these two losses, which is 1. 
Our flow loss $\mathcal{L}_{flow}$ is the $L2$ loss between warped frames. Thus, we set the weight of this loss to the same as the previous ones.
As for our temporal consistency loss $\mathcal{L}_{temp}$, we first set the weight of this L2 loss 	$\lambda_{t}$ to 1. However, we found that the generated results tend to be black and white. Then, we changed it to 0.1. Yet, this small weight makes this loss too weak to maintain the temporal consistency. Finally, we set it to 0.5, and obtained the current good results.

\begin{figure*}[t!]
	\begin{center}
		\captionsetup[subfigure]{labelformat=empty}
		\subfloat[Input Image]{\includegraphics[width=0.125\textwidth]{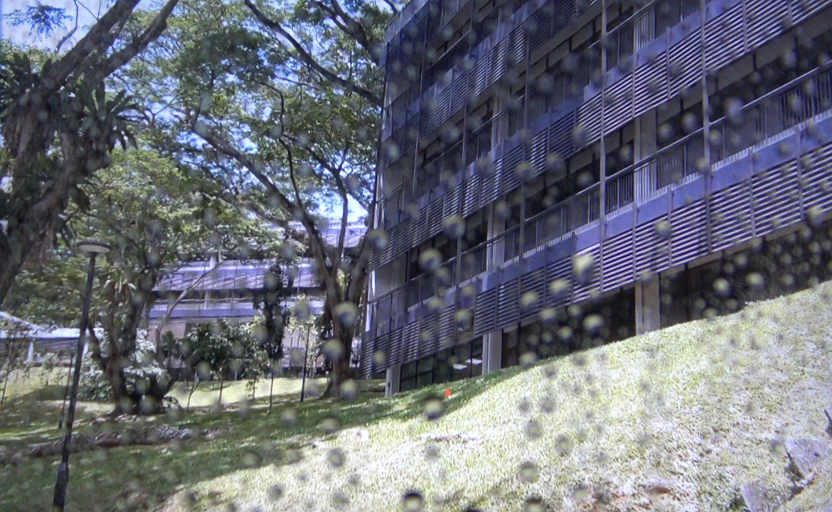}}\hfill	
		\subfloat[Groundtruth]{\includegraphics[width=0.125\textwidth]{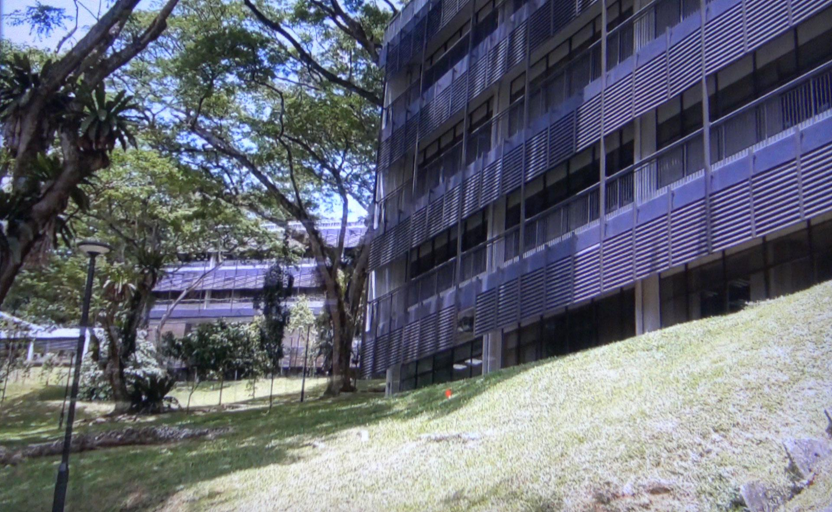}}\hfill	
		\subfloat[\textbf{Our Result} ]{\includegraphics[width=0.125\textwidth]{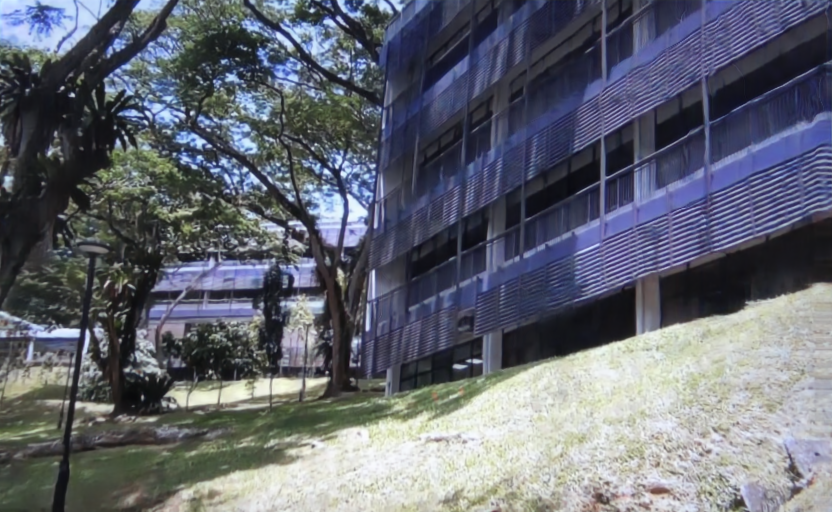}}\hfill
		\subfloat[Qian et al.~\cite{0Attentive} ]{\includegraphics[width=0.125\textwidth]{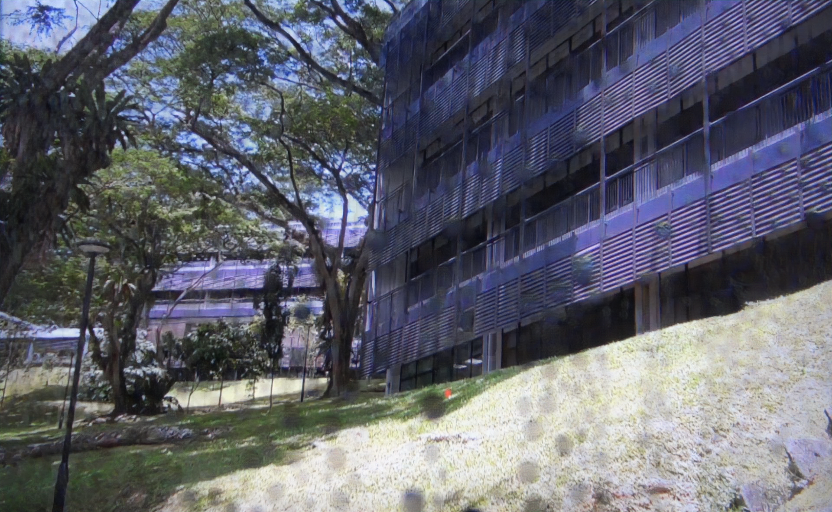}}\hfill
		\subfloat[Quan et al.~\cite{2019Deep} ]{\includegraphics[width=0.125\textwidth]{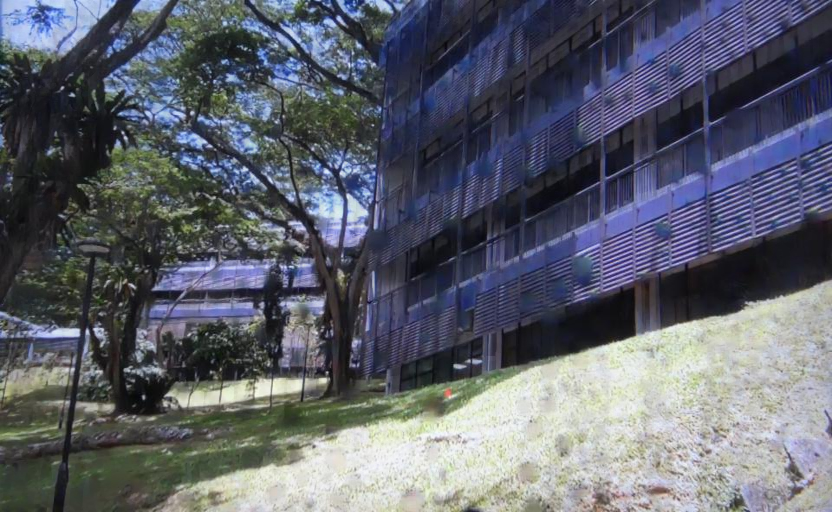}}\hfill
		\subfloat[Porav et al.~\cite{porav2019can}]{\includegraphics[width=0.125\textwidth]{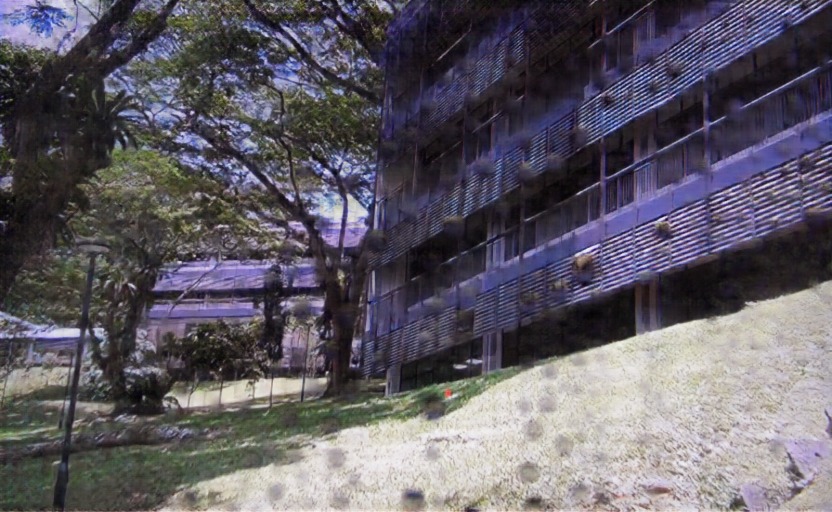}}\hfill
		\subfloat[Alletto et al.~\cite{Alletto2019AdherentRR} ]{\includegraphics[width=0.125\textwidth]{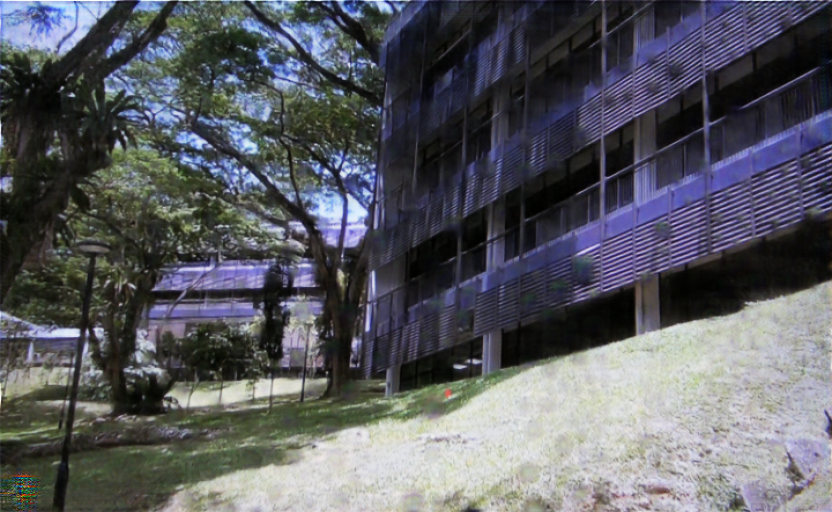}}\hfill
		\subfloat[CycleGAN~\cite{isola2017image} ]{\includegraphics[width=0.125\textwidth]{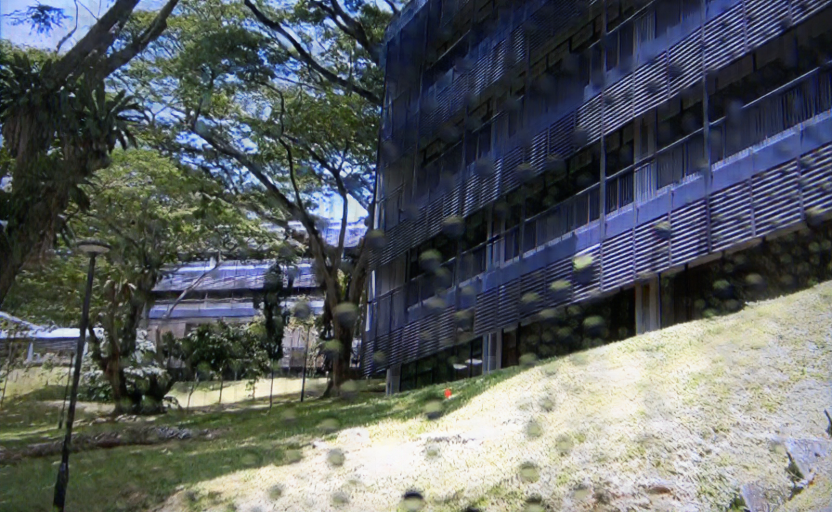}}\hfill\\
		
		\vspace{-0.1in}
		\subfloat[Input Image]{\includegraphics[width=0.125\textwidth]{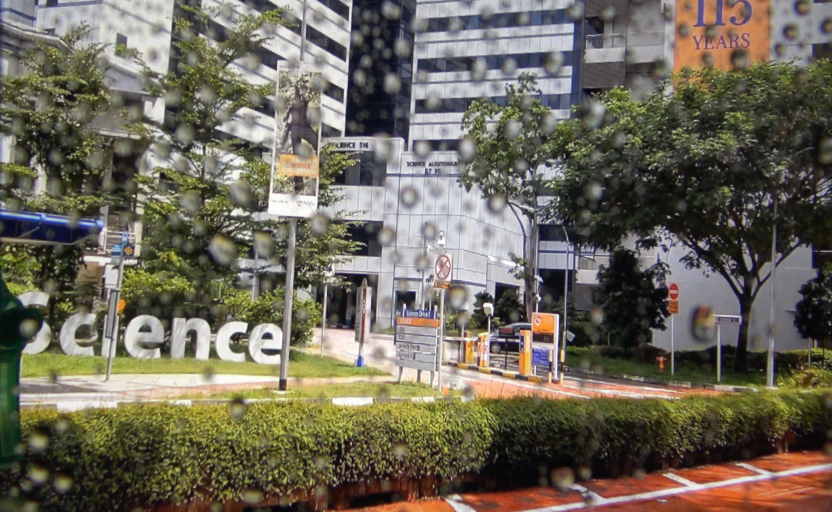}}\hfill	
		\subfloat[Groundtruth]{\includegraphics[width=0.125\textwidth]{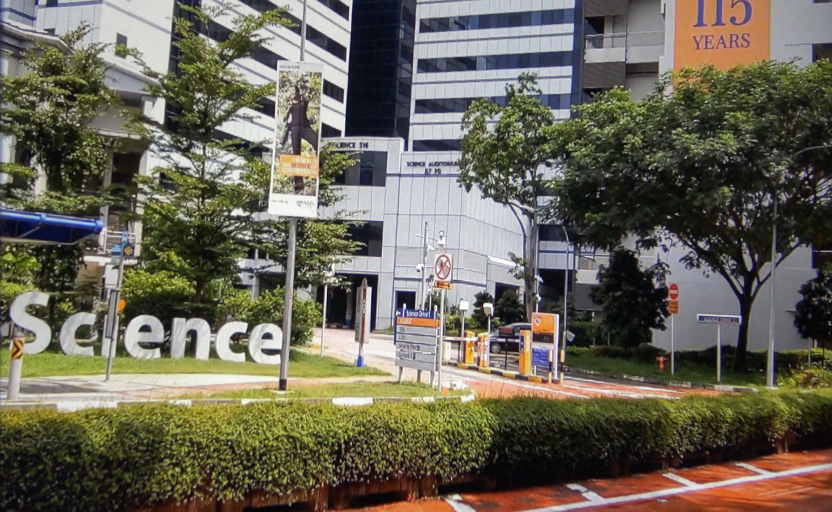}}\hfill	
		\subfloat[\textbf{Our Result} ]{\includegraphics[width=0.125\textwidth]{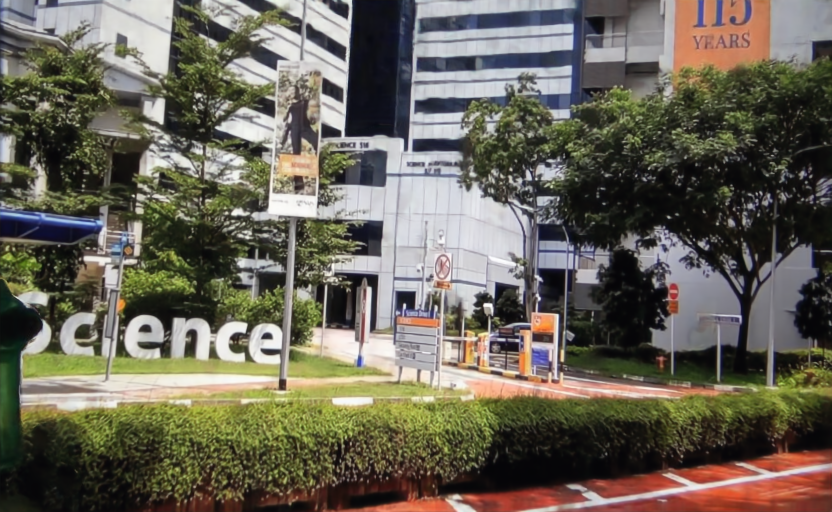}}\hfill
		\subfloat[Qian et al.~\cite{0Attentive} ]{\includegraphics[width=0.125\textwidth]{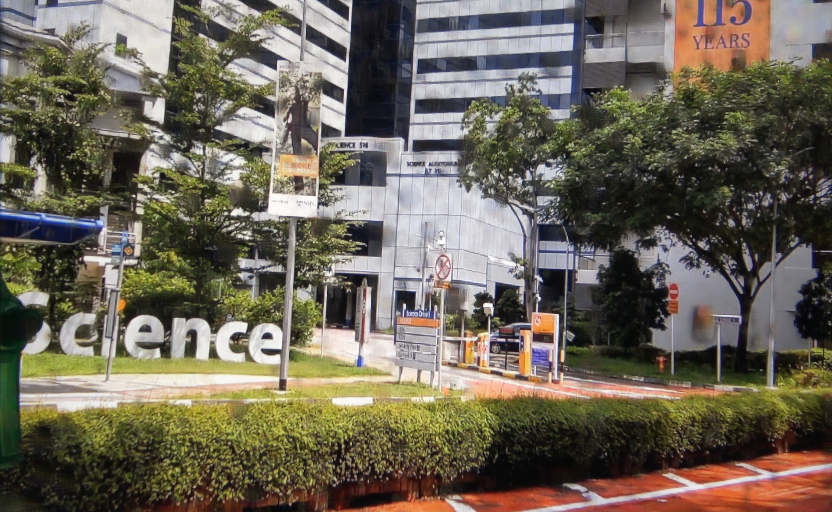}}\hfill
		\subfloat[Quan et al.~\cite{2019Deep} ]{\includegraphics[width=0.125\textwidth]{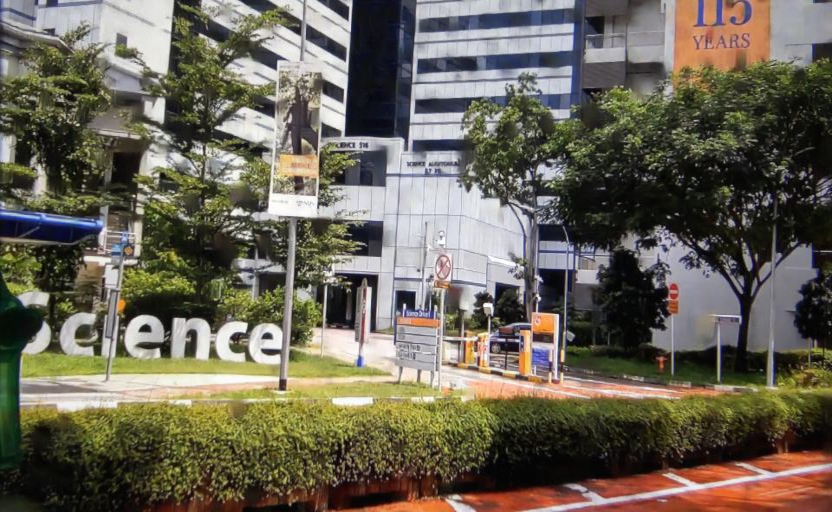}}\hfill
		\subfloat[Porav et al.~\cite{porav2019can}]{\includegraphics[width=0.125\textwidth]{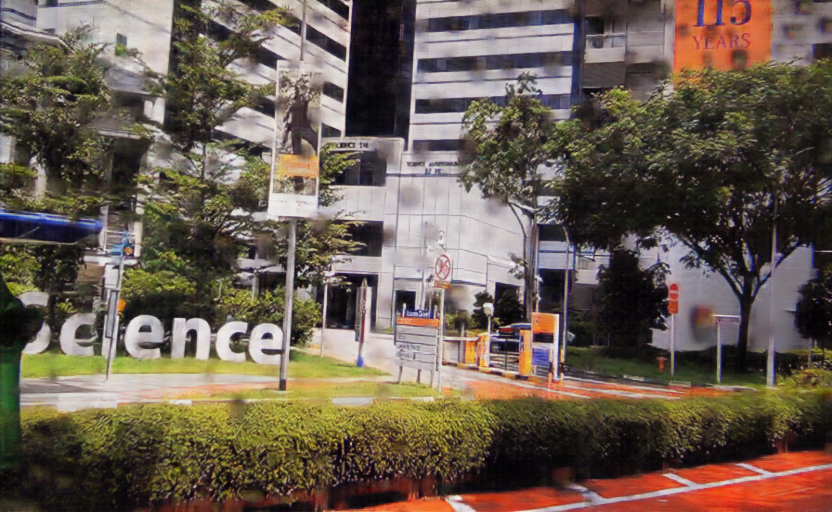}}\hfill
		\subfloat[Alletto et al.~\cite{Alletto2019AdherentRR}]{\includegraphics[width=0.125\textwidth]{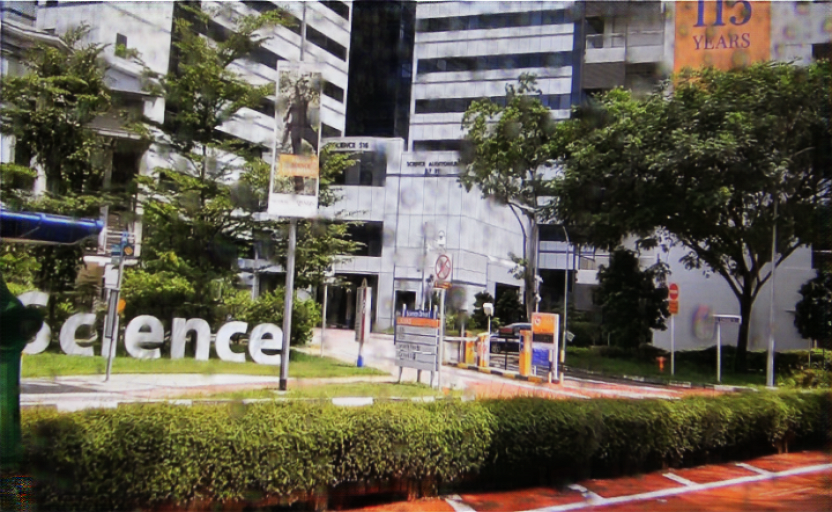}}\hfill
		\subfloat[CycleGAN~\cite{isola2017image} ]{\includegraphics[width=0.125\textwidth]{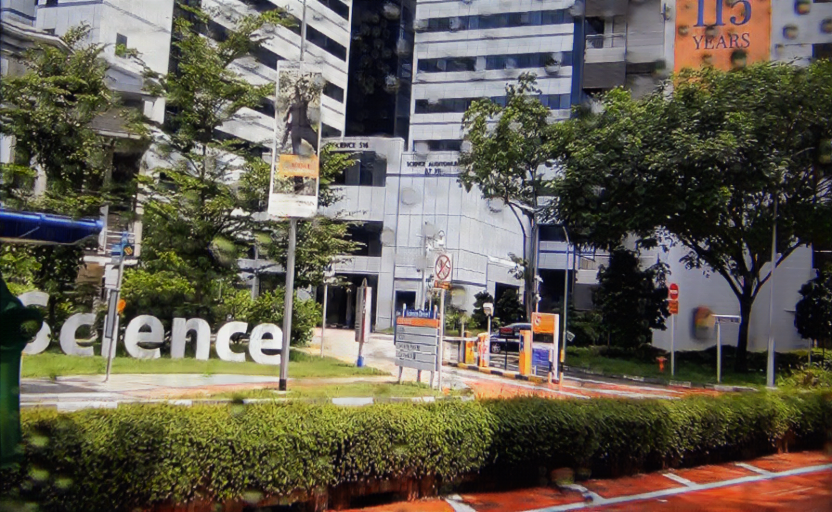}}\hfill\\
		
		\vspace{-0.1in}
		\subfloat[Input Image]{\includegraphics[width=0.125\textwidth]{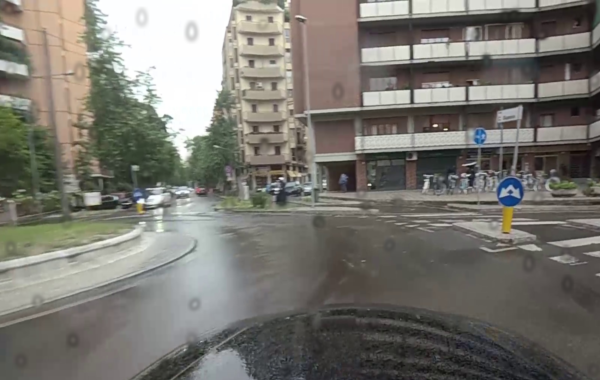}}\hfill	
		\subfloat[Groundtruth]{\includegraphics[width=0.125\textwidth]{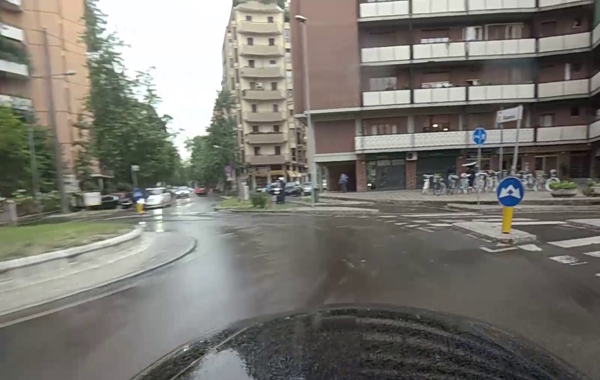}}\hfill	
		\subfloat[\textbf{Our Result} ]{\includegraphics[width=0.125\textwidth]{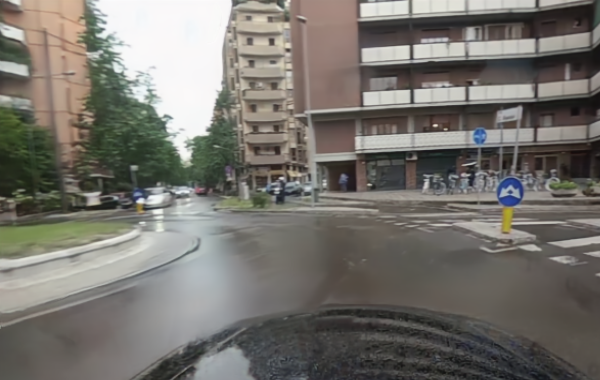}}\hfill
		\subfloat[Qian et al.~\cite{0Attentive} ]{\includegraphics[width=0.125\textwidth]{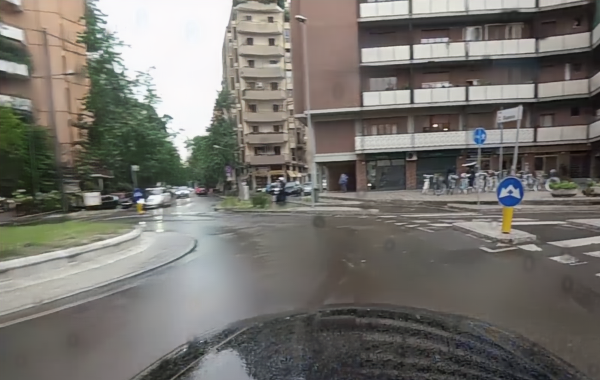}}\hfill
		\subfloat[Quan et al.~\cite{2019Deep} ]{\includegraphics[width=0.125\textwidth]{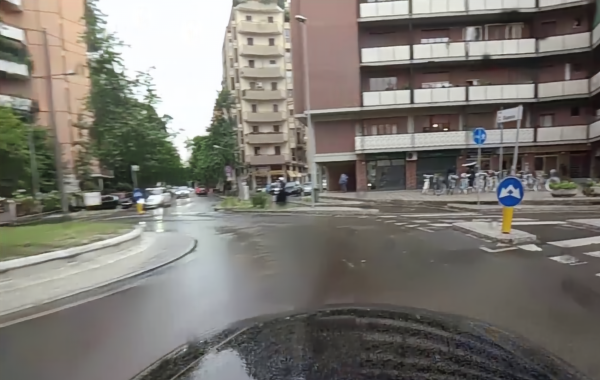}}\hfill
		\subfloat[Porav et al.~\cite{porav2019can}]{\includegraphics[width=0.125\textwidth]{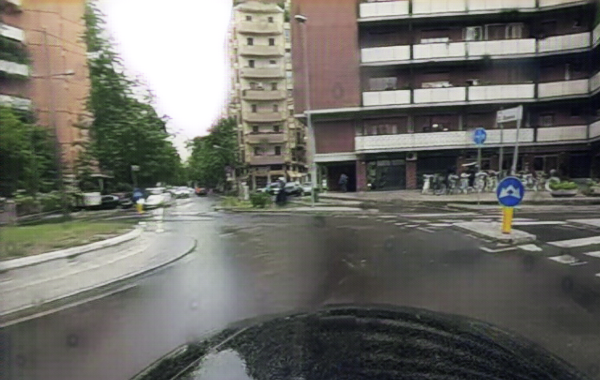}}\hfill
		\subfloat[Alletto et al.~\cite{Alletto2019AdherentRR}]{\includegraphics[width=0.125\textwidth]{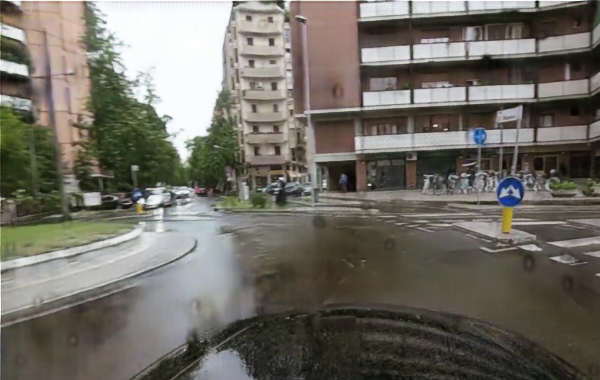}}\hfill
		\subfloat[CycleGAN~\cite{isola2017image} ]{\includegraphics[width=0.125\textwidth]{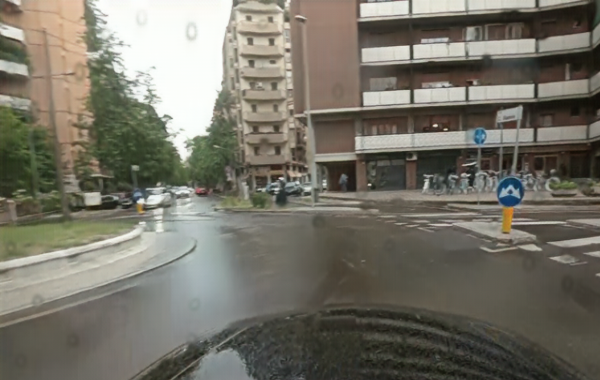}}\hfill\\
		
		\vspace{-0.1in}
		\subfloat[Input Image]{\includegraphics[width=0.125\textwidth]{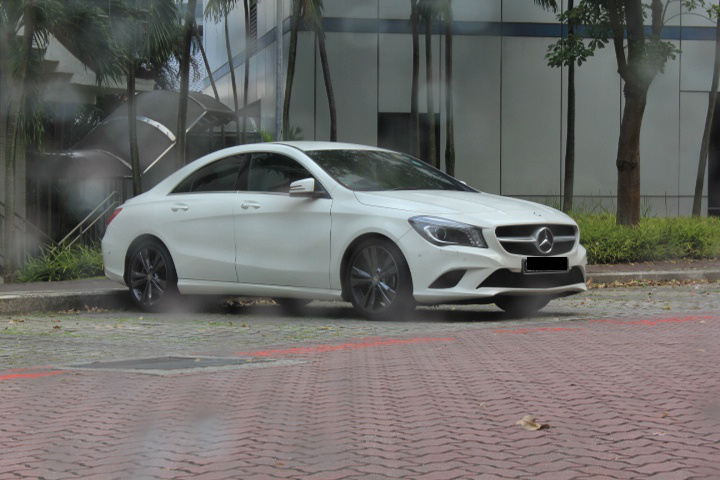}}\hfill	
		\subfloat[Groundtruth]{\includegraphics[width=0.125\textwidth]{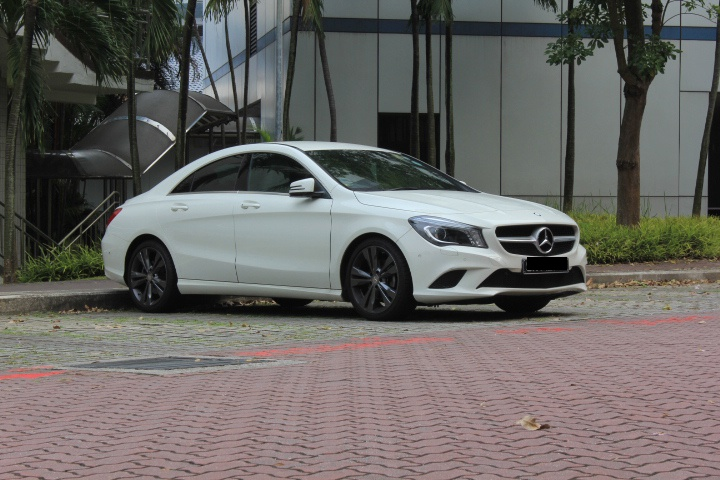}}\hfill	
		\subfloat[\textbf{Our First-Stage} ]{\includegraphics[width=0.125\textwidth]{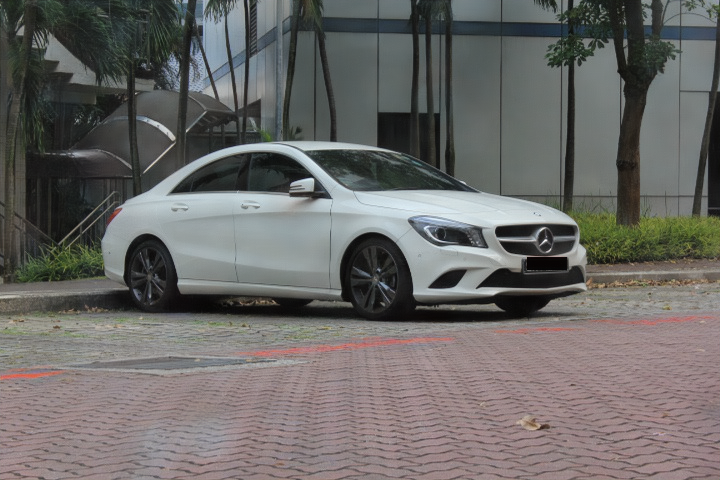}}\hfill
		\subfloat[Qian et al.~\cite{0Attentive} ]{\includegraphics[width=0.125\textwidth]{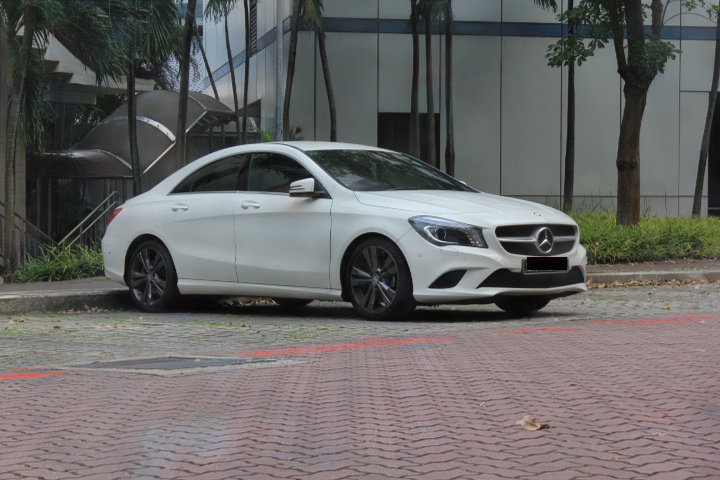}}\hfill
		\subfloat[Quan et al.~\cite{2019Deep} ]{\includegraphics[width=0.125\textwidth]{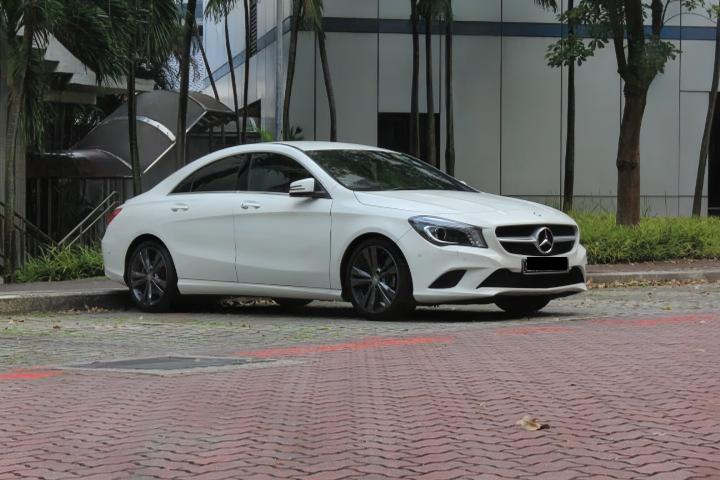}}\hfill
		\subfloat[Porav et al.~\cite{porav2019can}]{\includegraphics[width=0.125\textwidth]{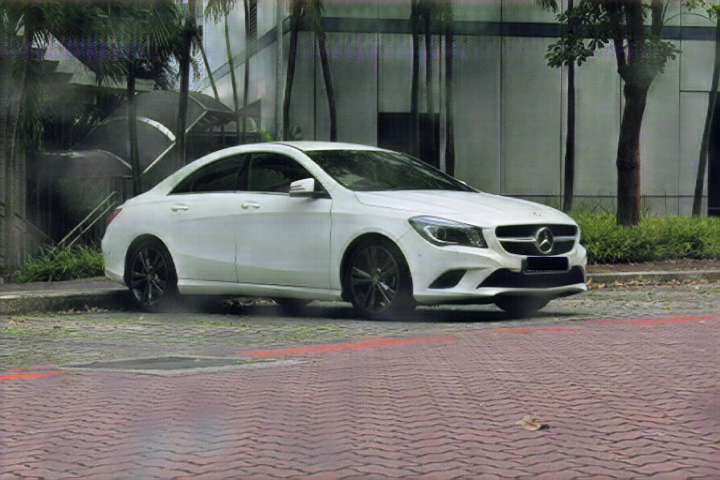}}\hfill
		\subfloat[Alletto et al.~\cite{Alletto2019AdherentRR}]{\includegraphics[width=0.125\textwidth]{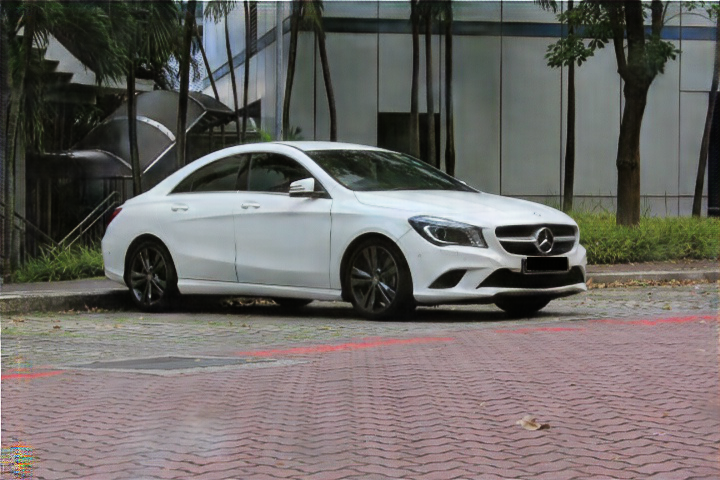}}\hfill
		\subfloat[CycleGAN~\cite{isola2017image} ]{\includegraphics[width=0.125\textwidth]{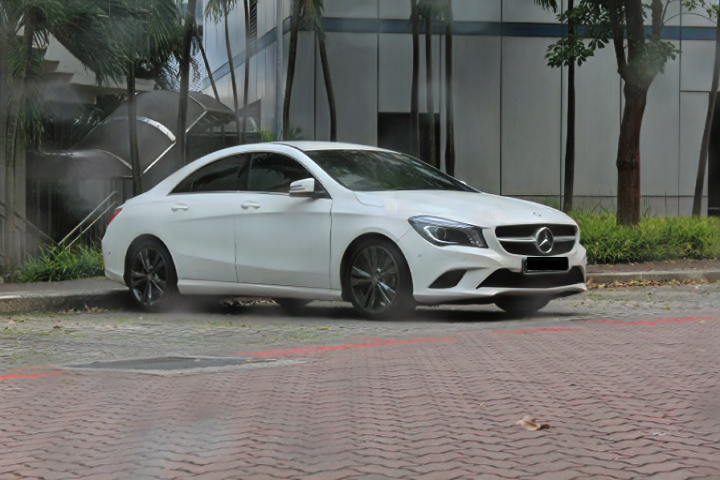}}\hfill\\
		
		\vspace{-0.1in}
		
	\end{center}
	\caption{Qualitative comparison of our method with the state-of-the-art methods and their variants on lab-recorded  raindrop images, the synthetic DRVE dataset and  the NUS dataset.}\label{synt1}
\end{figure*}

\begin{figure*}[t!]
	\begin{center}
		\captionsetup[subfigure]{labelformat=empty}
		\subfloat[Input Image]{\includegraphics[width=0.1428\textwidth]{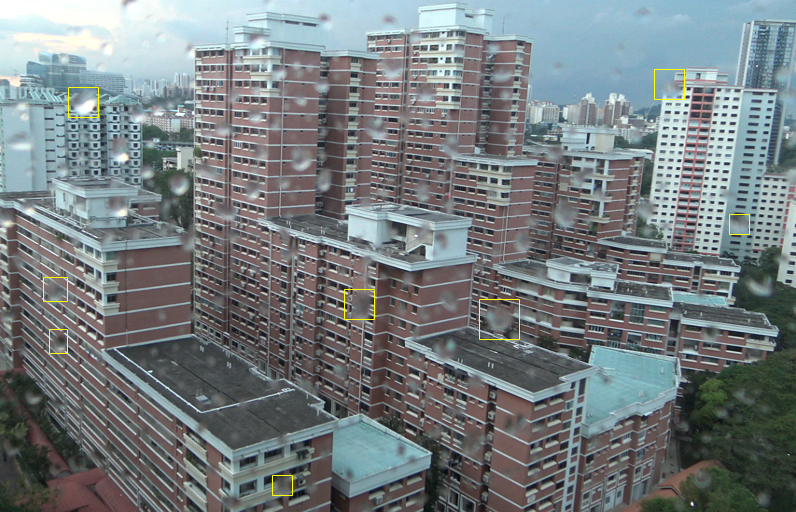}}\hfill	
		\subfloat[\textbf{Our Result}]{\includegraphics[width=0.1428\textwidth]{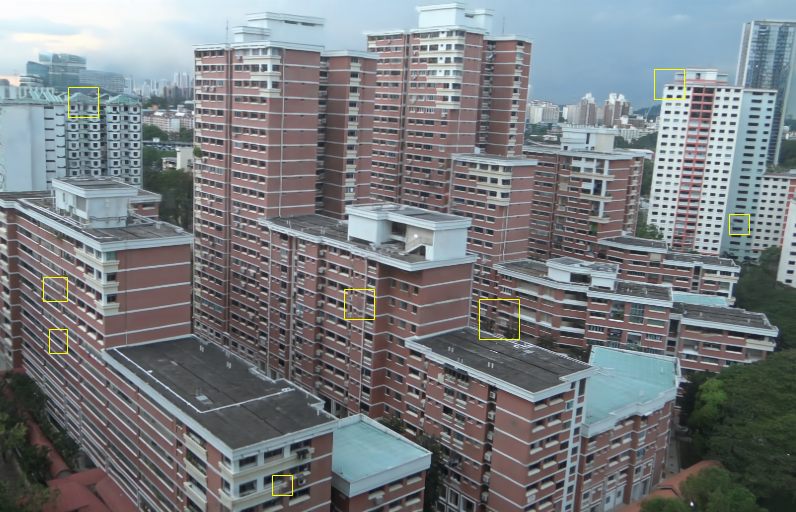}}\hfill
		\subfloat[Qian et al.~\cite{0Attentive}]{\includegraphics[width=0.1428\textwidth]{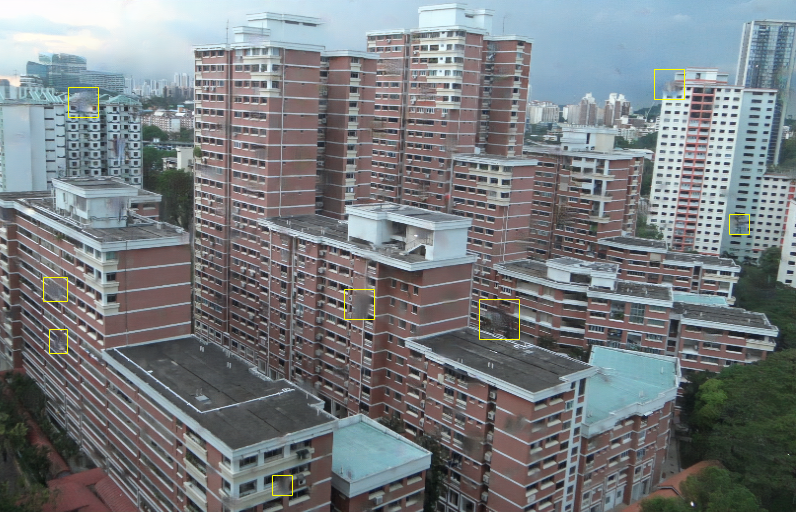}}\hfill
		\subfloat[Quan et al.~\cite{2019Deep}]{\includegraphics[width=0.1428\textwidth]{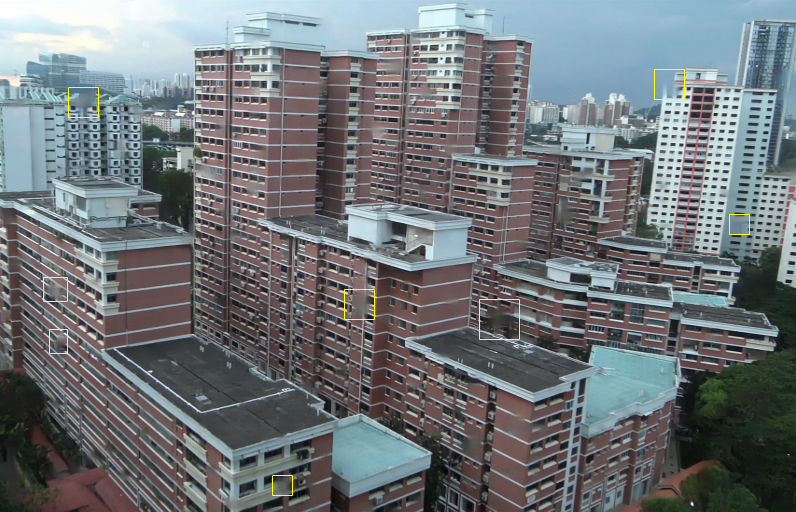}}\hfill
		\subfloat[Porav et al.~\cite{porav2019can}]{\includegraphics[width=0.1428\textwidth]{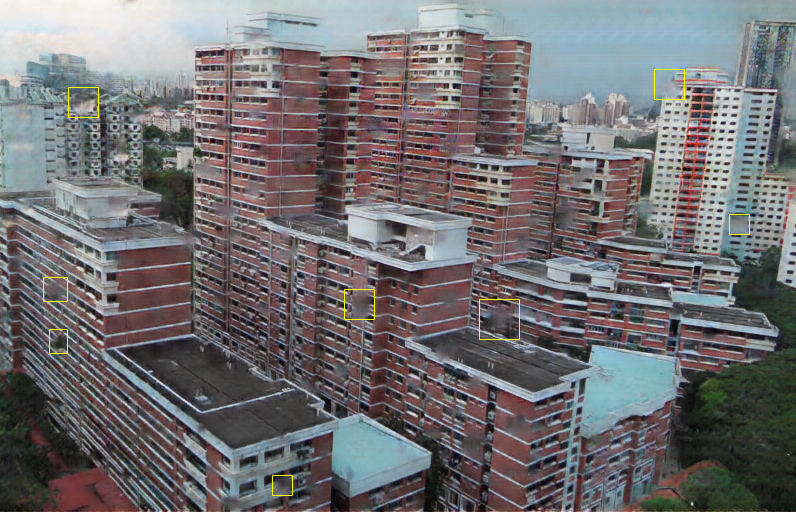}}\hfill
		\subfloat[Alletto et al.~\cite{Alletto2019AdherentRR}]{\includegraphics[width=0.1428\textwidth]{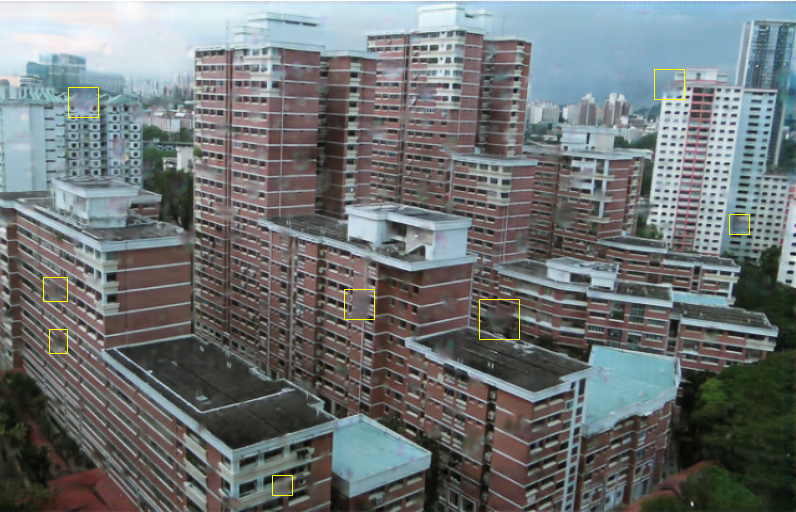}}\hfill
		\subfloat[CycleGAN~\cite{isola2017image} ]{\includegraphics[width=0.1428\textwidth]{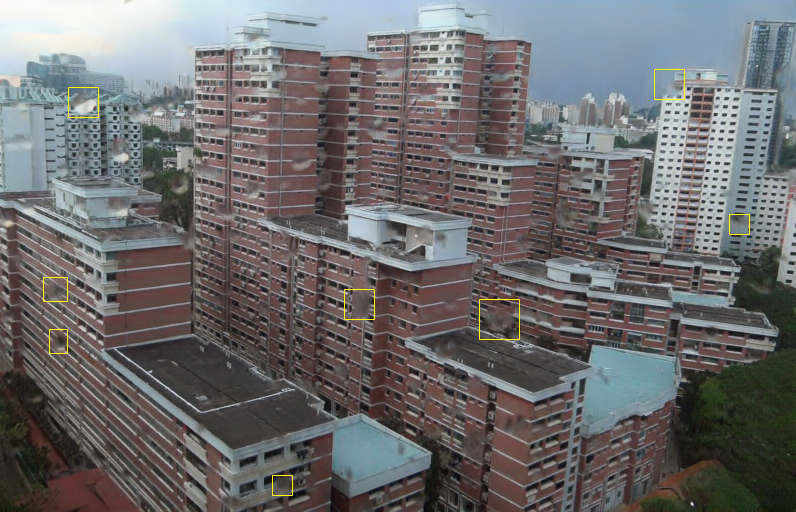}}\hfill\\	
		
		\vspace{-0.1in}
		\subfloat[Input Image]{\includegraphics[width=0.1428\textwidth]{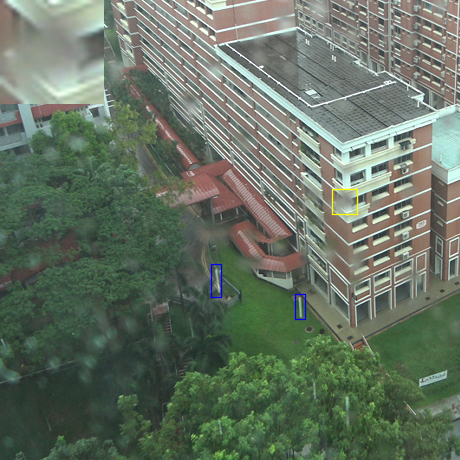}}\hfill	
		\subfloat[\textbf{Our Result} ]{\includegraphics[width=0.1428\textwidth]{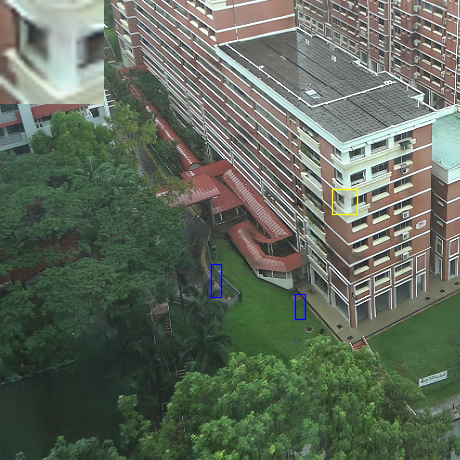}}\hfill
		\subfloat[Qian et al.~\cite{0Attentive} ]{\includegraphics[width=0.1428\textwidth]{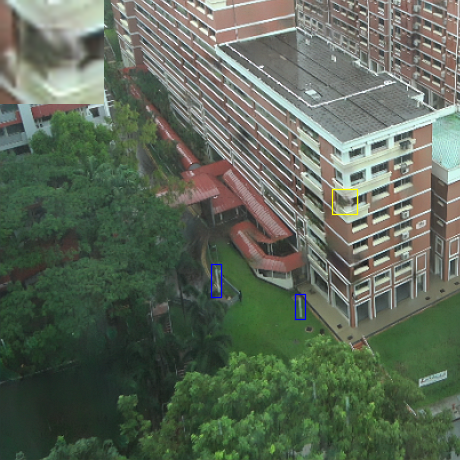}}\hfill
		\subfloat[Quan et al.~\cite{2019Deep} ]{\includegraphics[width=0.1428\textwidth]{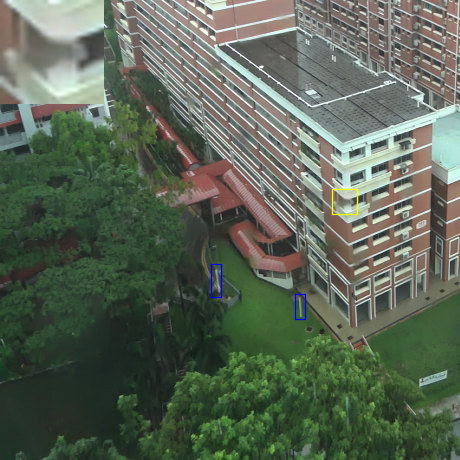}}\hfill
		\subfloat[Porav et al.~\cite{porav2019can}]{\includegraphics[width=0.1428\textwidth]{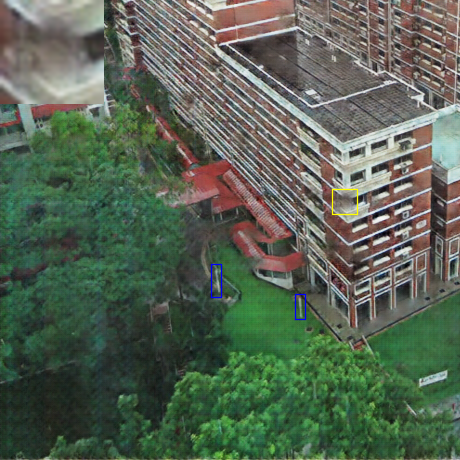}}\hfill
		\subfloat[Alletto et al.~\cite{Alletto2019AdherentRR}]{\includegraphics[width=0.1428\textwidth]{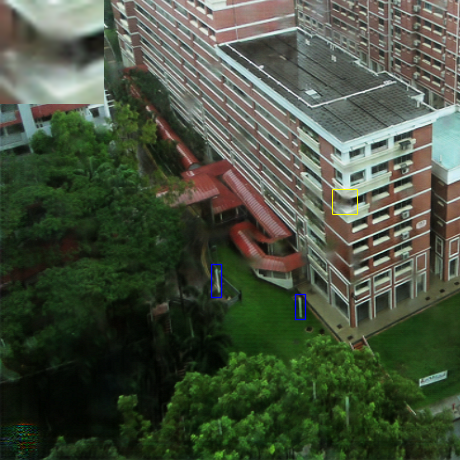}}\hfill
		\subfloat[CycleGAN~\cite{isola2017image} ]{\includegraphics[width=0.1428\textwidth]{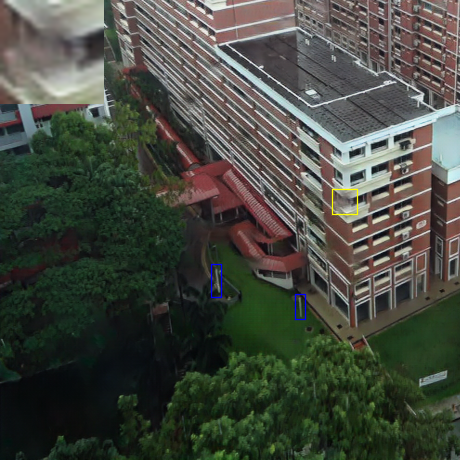}}\hfill\\
		
		\vspace{-0.1in}
		\subfloat[Input Image]{\includegraphics[width=0.1428\textwidth]{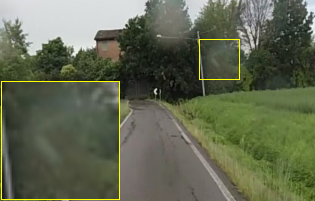}}\hfill	
		\subfloat[\textbf{Our Result} ]{\includegraphics[width=0.1428\textwidth]{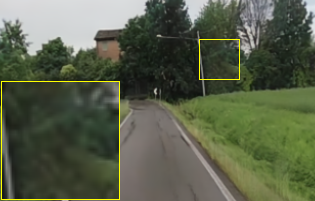}}\hfill
		\subfloat[Qian et al.~\cite{0Attentive} ]{\includegraphics[width=0.1428\textwidth]{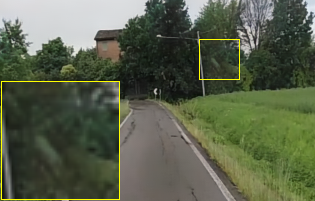}}\hfill
		\subfloat[Quan et al.~\cite{2019Deep} ]{\includegraphics[width=0.1428\textwidth]{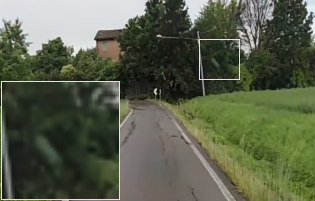}}\hfill
		\subfloat[Porav et al.~\cite{porav2019can}]{\includegraphics[width=0.1428\textwidth]{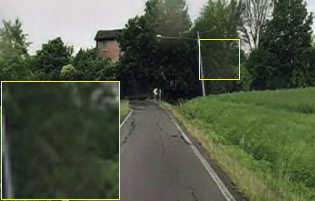}}\hfill
		\subfloat[Alletto et al.~\cite{Alletto2019AdherentRR}]{\includegraphics[width=0.1428\textwidth]{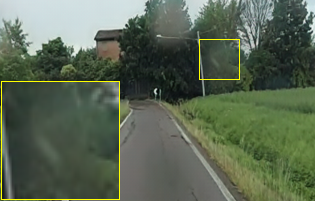}}\hfill
		\subfloat[CycleGAN~\cite{isola2017image} ]{\includegraphics[width=0.1428\textwidth]{img/cycle/DRVE.png}}\hfill\\
		
	\end{center}
	\caption{Qualitative comparison of our method with the state-of-the-art methods and their variants on real raindrop images and DR(eye)VE dataset. Zoom in the areas indicated by the yellow and blue boxes to see the effectiveness of our method. blue boxes indicate rain streaks.}\label{qual1}
\end{figure*}

\begin{table*}
	\small
	\centering
	
	\caption {Quantitative results on our lab-recorded raindrop data and synthetic DRVE dataset.} \label{table_quan}
	\begin{tabularx}{\textwidth}{ c|Y|Y|Y|Y|Y|Y|Y|Y }
		\hline
		\multirow{2}{*}{}   				& Video1-PSNR 	& Video1-SSIM		& Video2-PSNR	& Video2-SSIM		& Video3-PSNR	& Video3-SSIM		& DRVE-PSNR		& DRVE-SSIM\\
		\hline   
		\hline
		Input Image          		  		& 22.11 		& 0.8859 			& 22.90 		& 0.9120 			& 22.97			& 0.9282			& 36.03			& 0.9626\\
		\hline
		Qian et al.~\cite{0Attentive} 		& 23.41 		& 0.9053			& 25.39 		& 0.9488			& 25.33 		& 0.9575			& 40.35			& 0.9811\\
		\hline                    
		Quan et al.~\cite{2019Deep} 		& 23.49 		& 0.9035			& 25.89 		& 0.9490			& 25.53 		& 0.9567			& 38.77			& 0.9753\\
		\hline
		Porav et al.~\cite{porav2019can} 	& 22.87 		& 0.9003  			& 24.00 		& 0.9361			& 24.09 		& 0.9483			& 37.07			& 0.9492\\
		\hline
		Alletto et al.
		~\cite{Alletto2019AdherentRR} 	 	& 23.83 		& 0.8537 			& 23.96 		& 0.9007			& 23.81 		& 0.9010			& 37.27			& 0.9511\\
		\hline
		CycleGAN~\cite{isola2017image} 	 	& 22.37 		& 0.8854 			& 22.63 		& 0.9429			& 24.36 		& 0.9543			& 37.92			& 0.9675\\
		\hline
		Without Raindrop Mask		 	 	& 24.25 		& 0.9111 			& 29.19 		& 0.9707			& 27.04 		& 0.9625			& 39.13			& 0.9714\\
		\hline
		Without InitialNet			 	 	& 23.24 		& 0.9053 			& 23.79 		& 0.9202			& 24.05 		& 0.9456			& 40.83			& 0.9753\\
		\hline
		Without Alignment			 	 	& 23.01 		& 0.9042 			& 23.12 		& 0.9062			& 22.93 		& 0.9221			& 33.42			& 0.9247\\
		\hline
		Without Temporal Consistency 	 	& 23.96 		& 0.9107 			& 27.26 		& 0.9521			& 27.05 		& 0.9665			& 39.22			& 0.9714\\
		\hline
		\textbf{Our Result}  				&\textbf{24.51} &\textbf{0.9137}	&\textbf{30.35} &\textbf{0.9720}	&\textbf{27.42} &\textbf{0.9724}	&\textbf{42.44} &\textbf{0.9831}\\
	\end{tabularx} 
\end{table*}

\begin{figure*}
	\begin{center}
		\captionsetup[subfigure]{labelformat=empty}		
		\subfloat[Input Image]{\includegraphics[width=0.1666\textwidth]{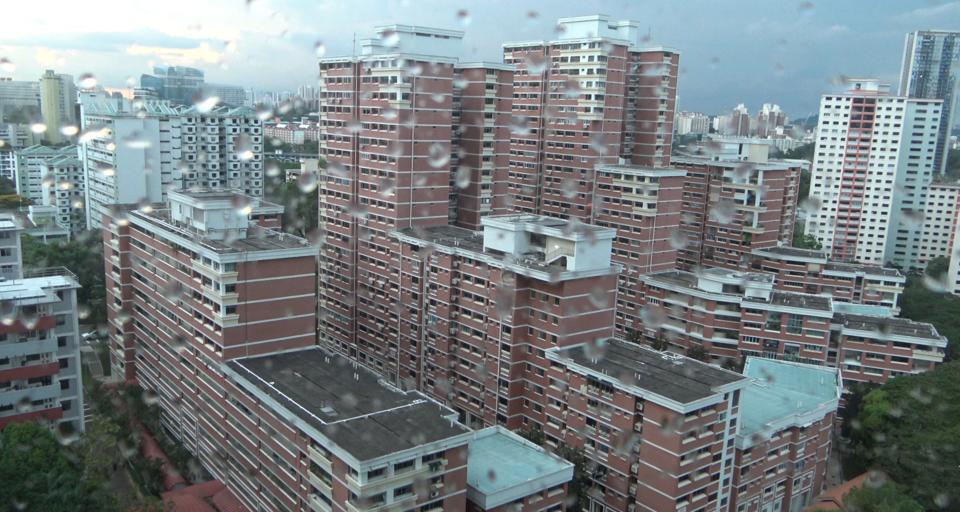}}\hfill	
		\subfloat[Full Module]{\includegraphics[width=0.1666\textwidth]{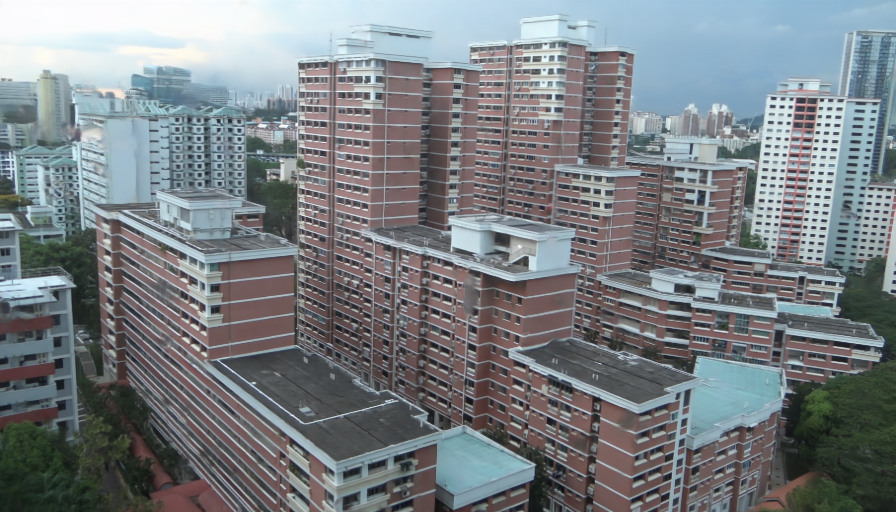}}\hfill
		\subfloat[Without Mask]{\includegraphics[width=0.1666\textwidth]{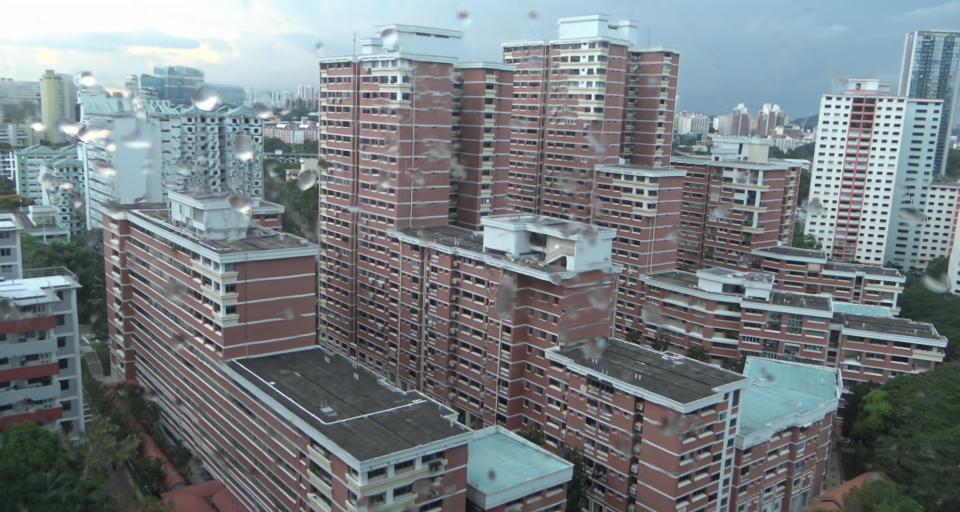}}\hfill
		\subfloat[Without InitialNet]{\includegraphics[width=0.1666\textwidth]{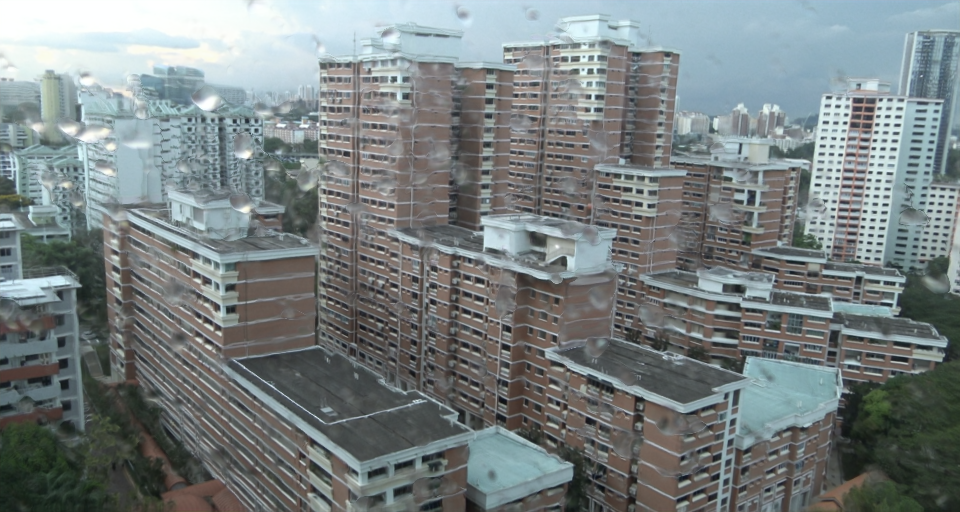}}\hfill
		\subfloat[Without Alignment]{\includegraphics[width=0.1666\textwidth]{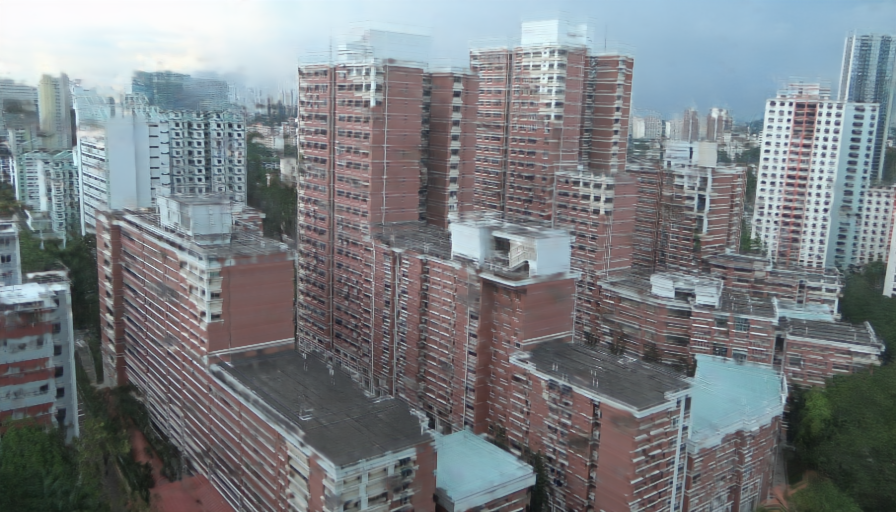}}\hfill
		\subfloat[Without Temporal Consistency]{\includegraphics[width=0.1666\textwidth]{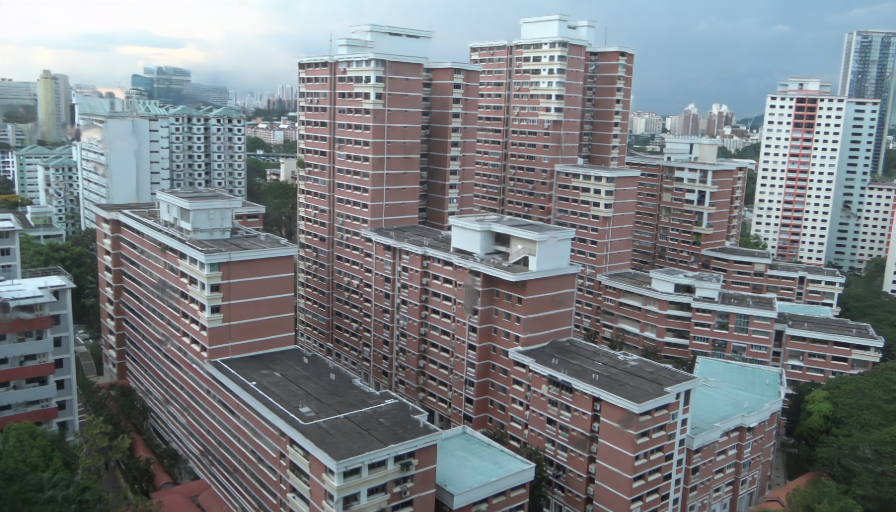}}\hfill\\
		
	\end{center}
	\caption{Ablation studies on raindrop mask, our InitialNet, alignment and temporal consistency.}\label{ablation}
\end{figure*} 

\section{Experimental Results}

\subsection{Data Collection and Implementation Details}

To train our single image raindrop removal model, we use the dataset from AttGAN~\cite{0Attentive}, which provides a raindrop image with a corresponding ground-truth clean image. We also use our lab-recorded raindrop data as the training data.

Our lab-recorded raindrop data is recorded using a monitor screen, a piece of glass (glass shield), and a camera. The idea is to record videos frame by frame through the glass shield. By using the glass shield with and without water sprayed on it, we can have video raindrop data with a ground-truth clean background. 
One additional advantage of such a strategy is that the light passing through the raindrop mainly comes from the entire scene from the monitor screen. This means what’s inside the raindrop is related to the displayed scene. By doing this, the generated raindrop images are closer to the real ones. 

We collected 33 raindrop videos with corresponding ground-truth clean frames by this strategy. Each video contains around 100 frames. We use 30 raindrop videos for training, and 3 videos for testing. The raindrop mask can be obtained by subtracting the clean image and raindrop image.
Then we collect real-world raindrop videos data from city roads.

The real-world raindrop videos are recorded in real-world rain scenes. By using the  glass shield in front of the camera that has raindrops attached to it, we can collect raindrop videos. The data is collected just like the way the raindrop image is formed in real application scenes.
Since the collection of the corresponding clean ground-truth images is intractable, these real-world raindrop videos are used for the qualitative evaluations (not for quantitative evaluations).
We totally collect 8 real-world raindrop videos and use 5 videos of them for training and 3 videos of them for testing.

During the training, videos are sampled and cropped into the size 512$\times$512 with a batch size 4.
Inspired by AttGAN~\cite{0Attentive}, we involve the raindrop mask as an attention mechanism in our single image raindrop removal model. For optical flow, a pretrained FlowNet2~\cite{ilg2017flownet} is used and finetuned with a learning rate 1e-7. Our video raindrop removal model is trained using Adam optimizer with a learning rate 1e-4.
The proposed raindrop removal method has two stages. In the first stage, around 3000 pairs of lab-collected data (from 30 videos) and 850 pairs from AttGAN~\cite{0Attentive} are used to train our single image raindrop removal model.  Our second stage is fully self-learned, thus we use 5 videos which is around 2000 real-world raindrop video frames for training. 

During the testing, 3 videos from lab-recorded raindrop data (around 100 frames for each video) and 3 videos with around 700 frames from real-world raindrop data are used.
Besides, we also use the DR(eye)VE ~\cite{palazzi2018predicting} dataset for qualitative evaluation that contains real-world videos with raindrops on the lens.
For quantitative evaluation, we need raindrop videos with corresponding ground-truth clean videos, which is not possible for real-world raindrop videos datasets. 
Therefore, we use the lab-recorded raindrop data with ground-truth clean backgrounds for quantitative evaluation.
Also, we add synthetic raindrops ~\cite{2021Raindrops} to clean videos of the DR(eye)VE~\cite{palazzi2018predicting} dataset for quantitative evaluation, which is called the synthetic DRVE.
In the quantitative evaluation, we take the average PSNR and SSIM of all frames in the input videos. Each input video contains around 100 frames.

\subsection{Comparison Results}
We choose five state-of-the-art raindrop removal methods as our baseline: Qian et al.~\cite{0Attentive}, Quan et al.~\cite{2019Deep}, Porav et al.~\cite{porav2019can}, Alletto et al. ~\cite{Alletto2019AdherentRR}, and CycleGAN~\cite{isola2017image}. 
We train the CycleGAN~\cite{isola2017image}, Porav et al.~\cite{porav2019can} and Alletto et al.~\cite{Alletto2019AdherentRR} methods with our lab-recorded data and AttGAN's~\cite{isola2017image} single image raindrop dataset.
Since most of these methods~\cite{0Attentive,2019Deep,porav2019can,isola2017image} only use a single image for training, extra video-based post-processing is used after each method. 

In our experiments, we use a state-of-the-art non-deep-learning video-based deraining method
Fastderain ~\cite{jiang2018fastderain} to refine the obtained background and further remove the raindrops as well as the rain streaks.
The reasons we use Fastderain ~\cite{jiang2018fastderain} as our post-processing method are as follows. 
Instead of adding Fastderain ~\cite{jiang2018fastderain} as the post-processing, there are three other options.
First, we can use the results of the single image method without any post-processing. However, it might be unfair to these methods, since video data can provide extra information which can be used for the enhancement.
Second, we might want to modify the architectures of these baselines, so that they can handle multiple frames as input.
However, modifying the architecture can hurt their performance. Hence, we didn’t do this.
Third, instead of non-deep-learning methods, we might want to use an existing deep-learning video-based deraining method as the post-processing

Third, instead of non-deep-learning methods, we might want to use an existing deep-learning video-based deraining method as the post-processing. However, compared to non-deep-learning methods, deep-learning methods have two issues. First, the performance of deep-learning methods is significantly dependent on the training data. If the training data is not suitable, the performance of deep-learning methods will not be optimum. Moreover, there may be a domain gap between the training and testing data, which may further reduce the performance. Second, the deep-learning methods may create some artifacts/fakeness/hallucination. As a result, this post-processing with a deep-learning method may hurt the baselines’ performance, which is unfair to these baselines.

Considering all these reasons above, we choose Fastderain ~\cite{jiang2018fastderain} as the postprocessing for the single-image baselines.
For the experiments on the paired data, PSNR and SSIM are used. 
Since we do not receive any response from the authors for the code-requiring email, we implement ~\cite{Alletto2019AdherentRR} by ourselves. The rest of these baseline methods are based on the codes released by the authors.

Quantitative evaluation of different methods is shown in Table.~\ref{table_quan}. Our results have better PSNR and SSIM than other state-of-the-art methods. The improvement mainly comes from better inpainting results, which prove the importance of utilizing temporal information.

We also compare our results with other methods qualitatively in Figs.~\ref{synt1} and \ref{qual1}. The test videos mainly come from our real-world raindrop data. 
In Fig.~\ref{synt1}, the first and second rows show the qualitative evaluation on our lab-recorded raindrop data, and the third row is the evaluation on the synthetic DRVE dataset.
Comparing our results to the ground-truth, our results have better raindrop removal performance than baseline methods.
The fourth row of Fig.~\ref{synt1} is evaluated on the DeRaindrop dataset which is a single image raindrop removal dataset. Thus, we compare only the outputs from our stage one results, the single image raindrop removal model, with other SOTA methods. 

Our single image raindrop removal model only aims to have initial results. 
As can be seen, our single image raindrop removal model has a promising raindrop removal result compared to other SOTA methods. 
Note that it is only an initial result and it will be refined in the multiple frame stage if the adjacent video frame data is available. 

As shown in the first row of Fig.~\ref{qual1}, other methods may suffer from artifacts especially around the background with rich details, like leaves or windows, while our method is able to reconstruct clean results.
The second row of Fig.~\ref{qual1} shows the testing results on video data with both raindrops and rain streaks. Our video raindrop removal model uses adjacent frames to predict the current frame that enables our method to remove both raindrops and rain streaks. However, other raindrop removal methods still have rain streaks left even with the post-processing model and suffer from poor removal results in some areas.

The third row  of Fig.~\ref{qual1} shows our method's results on the DR(eye)VE dataset. Note that there is a gray fake texture due to the raindrop in the yellow box. Our method identifies this fake texture from temporal information and successfully recovers the background tree's information from the adjacent frame, while all baseline methods keep this fake texture in their outputs.

\subsection{Ablation Studies}

To show the effectiveness of our raindrop mask, we remove the mask from the mask consistency loss and the mask correlation loss.
The third column of Fig.~\ref{ablation} shows the results trained with full-frame consistency and correlation losses (without mask). As can be seen, while the network is learning to recover texture details, raindrops are also recovered.

To prove that our InitialNet is effective, we directly feed the original inputs into our multiple frames raindrop removal module, showing the output in the fourth column of Fig.~\ref{ablation}. Without InitialNet, alignment will be misled by raindrops on inputs. From the results, some raindrops still remain. 
The fifth column of Fig.~\ref{ablation} shows the results without alignment (hence there is no optical flow and deformable convolution). 
Obviously, the results have blurry and overlapping issues due to mismatching. With the optical flow and deformable convolution, the features are aligned before feeding to the 3D-conv decoder. As a result, the output becomes clearer. 
In the last column of Fig.~\ref{ablation}, we show the output without temporal consistency loss. Since the output is not constrained by the temporal information, there are more artifacts and fakeness.  

\section{Conclusion}
In this paper, we have introduced a two-stage raindrop removal network, which is built based on temporal correlation in videos. The first stage takes this problem as single image raindrop removal. The model detects the locations of raindrops and provides initial results that can benefit the following process. In the second stage, aligned adjacent frames are used to predict a clean current frame free from both raindrops and rain streaks. The second stage only relies on temporal information for training that can be adjusted to different kinds of raindrop videos by self-learning. Our experiments demonstrate the effectiveness of our method.

\section*{Acknowledgment}
This work is supported by MOE2019-T2-1-130.

\bibliographystyle{IEEEtran}
\bibliography{IEEEabrv}
\vspace{-1cm}
\begin{IEEEbiography}[{\includegraphics[width=1in,height=1.25in,clip,keepaspectratio]{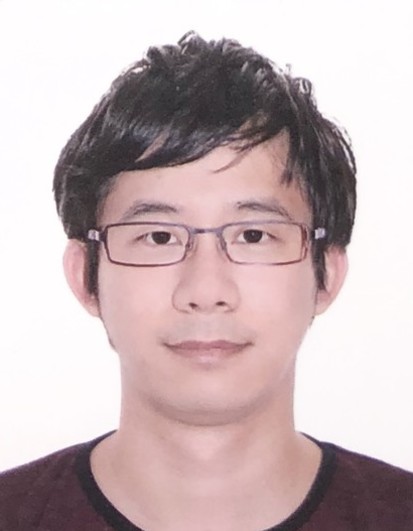}}]{Wending Yan}
	Wending Yan received his B.Eng. degree in School of Electrical and Electronic Engineering, Nanyang Technological University, Singapore in 2012. He is currently pursuing his Ph.D. degree in Computer Vision and Deep Learning at the Electrical and Computer Engineering Department, National University of Singapore, Singapore. His research interests include low-level vision, image restoration, particularly under bad weather.
\end{IEEEbiography}

\vspace{-1cm}

\begin{IEEEbiography}[{\includegraphics[width=1in,height=1.25in,clip,keepaspectratio]{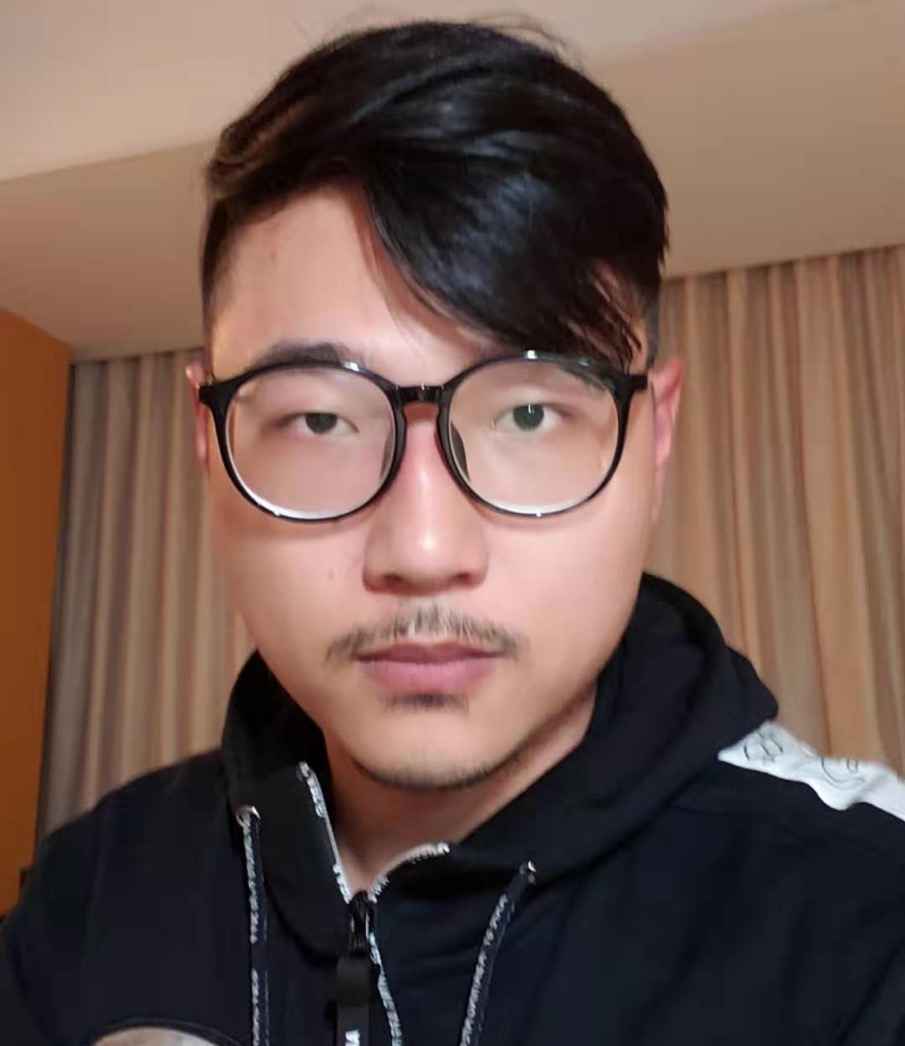}}]{Lu Xu}
	Lu Xu received a bachelor’s degree in digital media technology from Northeastern University, Shenyang, China. He is currently a Ph.D. candidate in college of information science and engineering from Northeastern University, Shenyang, China. His research interests include image /video enhancement and related vision problems..
\end{IEEEbiography}

\vspace{-1cm}

\begin{IEEEbiography}[{\includegraphics[width=1in,height=1.25in,clip,keepaspectratio]{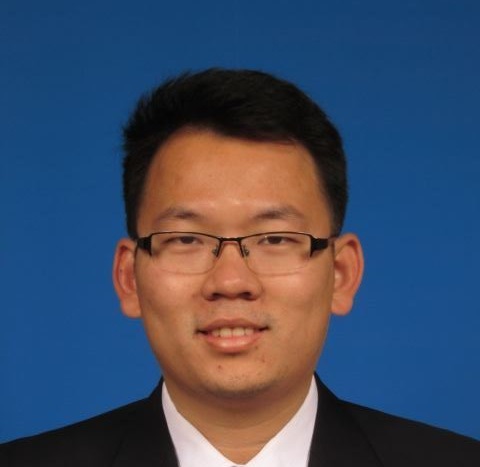}}]{Wenhan Yang}
	Wenhan Yang received the B.S degree and Ph.D. degree (Hons.) in computer science from Peking University, Beijing, China, in 2012 and 2018. He is currently a presidential postdoc fellow with the School of EEE, Nanyang Technological University, Singapore.
		He was a postdoctoral research fellow with the Department of Computer Science, City University of Hong Kong, during 2019-2021. 
		His current research interests include image/video processing/restoration, bad weather restoration, human-machine collaborative coding.
\end{IEEEbiography}

\vspace{-1cm}

\begin{IEEEbiography}[{\includegraphics[width=1in,height=1.25in,clip,keepaspectratio]{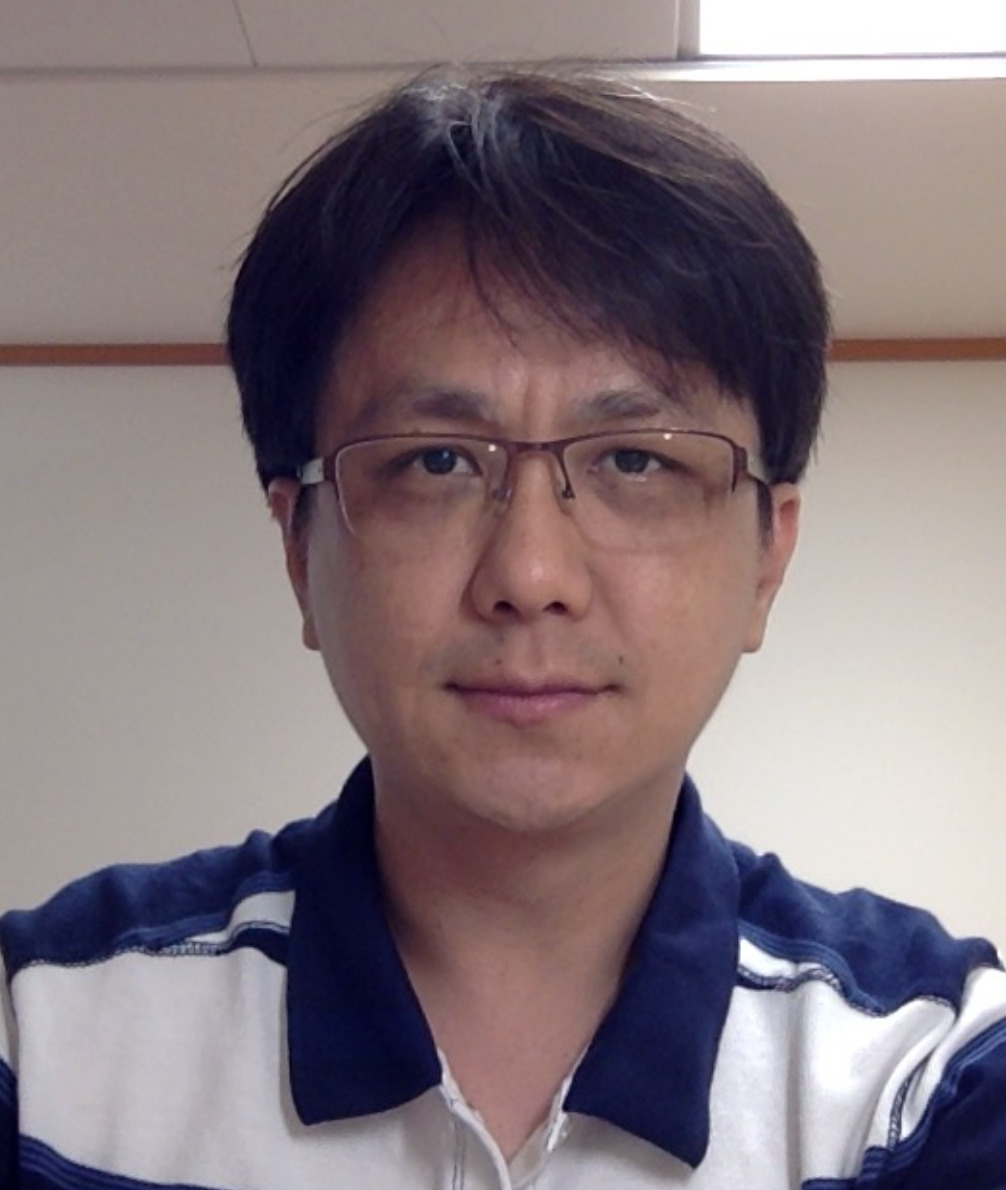}}]{Robby T. Tan}
	Robby T. Tan received the PhD degree in computer science from the University of Tokyo. 
	He is now an associate  professor at both Yale-NUS College and ECE (Electrical and Computing Engineering), National University of Singapore.
	Previously, he was an assistant professor at Utrecht University. 
	His research interests include computer vision and deep learning, particularly in the domains of low level vision (bad weather/nighttime, color analysis, physics-based vision, optical flow, etc.), human pose/motion analysis, and applications of deep learning in healthcare. 
	He is a member of the IEEE.
\end{IEEEbiography}




\end{document}